\documentclass[11pt]{article}

\usepackage[final]{acl}

\usepackage{times}
\usepackage{latexsym}

\usepackage[utf8]{inputenc} 
\usepackage[T1]{fontenc}    
\usepackage{hyperref}       
\usepackage{url}            
\usepackage{booktabs}       
\usepackage{amsfonts}       
\usepackage{nicefrac}       
\usepackage{microtype}      
\usepackage{xcolor}         
\usepackage{pifont}
\usepackage{bbding}
\usepackage{lineno}

\usepackage{multirow}
\usepackage{amsmath}
\usepackage{booktabs}

\definecolor{darkblack}{rgb}{0, 0, 0.5}
\hypersetup{colorlinks=true, citecolor=darkblue, linkcolor=darkblue, urlcolor=darkblack}

\usepackage{algorithm}
\usepackage{algpseudocode}
\usepackage{graphicx}
\usepackage{makecell}
\usepackage{fontawesome}
\usepackage{subcaption}  
\RequirePackage{subcaption}
\usepackage{wasysym}
\usepackage{utfsym}
\usepackage{fontawesome}
\usepackage[T1]{fontenc}

\usepackage[utf8]{inputenc}

\usepackage{microtype}

\usepackage{inconsolata}

\usepackage{graphicx}

\title{Nirvana: A Specialized Generalist Model With Task-Aware Memory Mechanism}

\author{
Yuhua Jiang$^{1,2,3}$,
Shuang Cheng$^{1,4}$,
Yihao Liu$^{1,2}$,
Ermo Hua$^{1,2}$,
Che Jiang$^{1,2}$,\\ 
{\bfseries\color{black}
Weigao Sun$^{1}$,
Yu Cheng$^{1}$,
Feifei Gao$^{2}$,
Biqing Qi$^{1}$\thanks{Corresponding author.},
Bowen Zhou$^{1,2}$ }\\
$^{1}$Shanghai AI Laboratory 
$^{2}$Tsinghua University \\ 
$^{3}$Xiong'an Anying Technology Co., Ltd. 
$^{4}$Zhejiang University 
}

\begin{document}
\maketitle
\begin{abstract}
Large Language Models (LLMs) excel at general language tasks but struggle in specialized domains. Specialized Generalist Models (SGMs) address this by preserving broad capabilities while adapting to target domains. However, existing architectures provide limited support for task-guided specialized memory mechanisms.
In this work, we introduce Nirvana, an SGM featuring specialized memory, linear-time complexity, and test-time task information extraction. Central to Nirvana are:
(1) Task-Aware Memory Trigger ($\textit{Trigger}$), which treats each input as a self-supervised fine-tuning task and adjusts task-related parameters on the fly; and
(2) Specialized Memory Updater ($\textit{Updater}$), which dynamically consolidates task-relevant context.
Nirvana matches or surpasses LLM baselines on general benchmarks and achieves the lowest perplexity across specialized domains including biomedicine, finance, and law. On the challenging task of Magnetic Resonance Imaging (MRI), we attach lightweight codecs to the frozen Nirvana backbone and fine-tune them on paired k-space signals and images. Nirvana achieves higher-fidelity reconstructions than conventional LLM-based models, with Trigger providing effective domain-specific adaptation.
Ablation studies confirm that removing Trigger leads to substantial degradation across all tasks, underscoring its essential role in task-aware specialization.
Models are available at \url{https://huggingface.co/collections/YuhuaJiang/nirvana}.
Code is available at \url{https://github.com/YuhuaJiang2002/Nirvana}.
\end{abstract}

\section{Introduction}

\begin{table*}[ht]
  \centering
    \resizebox{15cm}{!}{%
\begin{tabular}{llllllll} 
\hline Model &  \makecell[l]{Dynamic\\ Decay}  & \makecell[l]{Non-\\Linearity} & \makecell[l]{Local\\ Optimum} & \makecell[l]{Specialized\\ Memory} & \makecell[l]{Memory Update\\ Operation} \\
\hline Attention  & $\usym{2717}$ & $\checkmark$ & $\checkmark$ & $\usym{2717}$ & $\boldsymbol{M}_t=\boldsymbol{M}_{t-1} \cup\left\{\left(\boldsymbol{k}_t, \boldsymbol{v}_t\right)\right\}$ \\
\hline SWA & $\usym{2717}$ & $\checkmark$ & $\checkmark$ & $\usym{2717}$ & $\boldsymbol{M}_t=\left(\boldsymbol{M}_{t-1} \backslash\left\{\left(\boldsymbol{k}_c, \boldsymbol{v}_c\right)\right\}\right) \cup\left\{\left(\boldsymbol{k}_t, \boldsymbol{v}_t\right)\right\}$ \\
\hline Naive Linear Attention & $\usym{2717}$ & $\usym{2717}$ & $\usym{2717}$ & $\usym{2717}$ & $\boldsymbol{M}_t=\boldsymbol{M}_{t-1}+\boldsymbol{v}_t \boldsymbol{k}_t^{\top}$ \\
\hline DeltaNet  & $\usym{2717}$ & $\usym{2717}$ & $\usym{2717}$ &  $\usym{2717}$ & $\boldsymbol{M}_t=\left(\boldsymbol{I}-\beta_t \boldsymbol{k}_t \boldsymbol{k}_t^{\top}\right) \boldsymbol{M}_{t-1}+\beta_t \boldsymbol{v}_t \boldsymbol{k}_t^{\top}$ \\
\hline Longhorn & $\usym{2717}$ & $\usym{2717}$ & $\usym{2717}$ & $\usym{2717}$ & $\boldsymbol{M}_t=\left(\boldsymbol{I}-\delta_t \boldsymbol{k}_t \boldsymbol{k}_t^{\top}\right) \boldsymbol{M}_{t-1}+\delta_t  \boldsymbol{v}_t \boldsymbol{k}_t^{\top}$ \\ 
\hline RetNet/Lightning  & $\usym{2717}$ & $\usym{2717}$ & $\usym{2717}$ & $\usym{2717}$ & $\boldsymbol{M}_t=\alpha \boldsymbol{M}_{t-1}+\boldsymbol{v}_t \boldsymbol{k}_t^{\top}$ \\
\hline GLA  & $\checkmark$ & $\usym{2717}$ & $\usym{2717}$ & $\usym{2717}$ & $\boldsymbol{M}_t=\operatorname{Diag}\left(\alpha_t\right) \boldsymbol{M}_{t-1}+\boldsymbol{v}_t \boldsymbol{k}_t^{\top}$ \\
\hline HGRN2  & $\checkmark$ & $\usym{2717}$ & $\usym{2717}$ & $\usym{2717}$ & $\boldsymbol{M}_t=\operatorname{Diag}\left(\boldsymbol{a}_t\right) \boldsymbol{M}_{t-1}+\boldsymbol{v}_t (\boldsymbol{1}-\boldsymbol{a}_t)^{\top}$ \\
\hline Mamba2  & $\checkmark$ & $\usym{2717}$ & $\usym{2717}$ & $\usym{2717}$ & $\boldsymbol{M}_t=\alpha_t \boldsymbol{M}_{t-1}+ \beta_t \boldsymbol{v}_t \boldsymbol{k}_t^{\top}$ \\
\hline PolySketchFormer & $\usym{2717}$ & $\checkmark$ & $\usym{2717}$ & $\usym{2717}$ & $\boldsymbol{M}_t=\boldsymbol{M}_{t-1}+\boldsymbol{v}_t\left(\boldsymbol{k}_t^{\top}\right)^p$ \\
\hline TTT & $\usym{2717}$ & $\checkmark$ & $\usym{2717}$ & $\usym{2717}$ & $\boldsymbol{M}_t=\boldsymbol{M}_{t-1}-\eta_t \nabla \ell\left(\boldsymbol{M}_{t-1} (\boldsymbol{k}_t), \boldsymbol{v}_t\right)$ \\
\hline RWKV-7 & $\checkmark$ & $\usym{2717}$ & $\usym{2717}$ & $\usym{2717}$ & $\boldsymbol{M}_t=\left(\operatorname{Diag}\left(\alpha_t\right)-\beta_t \boldsymbol{k}_t \boldsymbol{k}_t^{\top}\right) \boldsymbol{M}_{t-1}+\beta_t \boldsymbol{v}_t \boldsymbol{k}_t^{\top}$ \\
\hline Gated DeltaNet &  $\checkmark$ & $\usym{2717}$ & $\usym{2717}$ & $\usym{2717}$ & $\boldsymbol{M}_t=\alpha_t\left(\boldsymbol{I}-\beta_t \boldsymbol{k}_t \boldsymbol{k}_t^{\top}\right) \boldsymbol{M}_{t-1}+\beta_t \boldsymbol{v}_t \boldsymbol{k}_t^{\top}$ \\
\hline Titans/Miras & $\checkmark$ & $\checkmark$ & $\usym{2717}$ & $\usym{2717}$ & 
\makecell[l]{
$\boldsymbol{M}_t=\alpha_t \boldsymbol{M}_{t-1}+\boldsymbol{S}_t$ \\
$\boldsymbol{S}_t=\eta_t \boldsymbol{S}_{t-1}-\eta_t \nabla \ell\left(\boldsymbol{M}_{t-1} ( \boldsymbol{k}_t), \boldsymbol{v}_t\right)$
} \\
\hline Nirvana & $\checkmark$ & $\checkmark$ & $\checkmark$ & $\checkmark$ &       \makecell[l]{
        $\boldsymbol{M}_t$ = $\gamma_t\{\alpha_t(\boldsymbol{I}-\beta_t \boldsymbol{k}_t \boldsymbol{k}_t^\top)\,\boldsymbol{M}_{t-1}^{\text{LA}}+ \beta_t \boldsymbol{v}_t \boldsymbol{k}_t^\top\}$ \\
        $\displaystyle \cup \,\eta_t\{\bigl(\boldsymbol{M}_{t-1}^{\text{SWA}}\setminus\{(\boldsymbol{k}_c,\boldsymbol{v}_c)\}\bigr)\cup(\boldsymbol{k}_t,\boldsymbol{v}_t)\}$
      } \\
\hline
\end{tabular}
   }
  \caption{
A summary of some modern LLM architectures. We compare them based on 4 characteristics: Dynamic Decay: adaptively forget memory about the past; Non-Linearity: beyond linear algebra operations such as matrix multiplications; Local Optimum: extract the second-order information about tokens; Specialized Memory: adaptively memorize the context according to the task information. 
}
\label{summary}
\end{table*}

Large Language Models (LLMs) have significantly advanced general language processing, but still have limitations in specialized tasks \citep{s1,s2,s3,s4}. 
For instance, while an LLM can describe the rules of the game of Go, it struggles to match the deep, domain-specific strategic reasoning of expert Go programs like AlphaGo.
To solve this problem, Specialized Generalist Models (SGMs) \citep{zhang2024sgi} are proposed to retain broad, generalist capabilities while achieving expert-level performance in at least one (and ideally multiple) specialized domains. 
SGMs play a pivotal role in real deployments, e.g., medicine and other safety-critical workflows, which demand both general reasoning ability and domain-expert inference accuracy, together with verifiable use of external knowledge and tools \citep{wang2025adapt,lewis2020rag,schick2023toolformer,yao2023react}. 


Specifically, the specialized memory mechanism of SGMs requires that models can identify the task information on the fly and then adapt their internal pathways and memory use, explicit retrieval and non-parametric memory \citep{fedus2021switch,lepikhin2020gshard,jiang2024mixtral, lewis2020rag,borgeaud2022retro,khandelwal2019knnlm,wu2022memorizing,behrouz2025s}, as well as the ability to dynamically choose the methodology to memorize.
Diverse memory mechanisms have been explored to capture, store, and adapt contextual information, which is summarized in Table~\ref{summary}. 
However, existing LLM architectures still exhibit limitations in supporting flexible and specialized memory mechanisms, and it remains an open question \textbf{how memory mechanism can be adaptively adjusted in a task-specific manner during test time.}

To answer this question, we propose a novel SGM called Nirvana, which realizes specialized and non-linear memory mechanism with dynamic decay and second-order information. 
We propose a Task-Aware Memory Trigger ($\textit{Trigger}$), which enables dynamic self-supervised fine-tuning to adapt to domain shifts. 
By turning each incoming sample into a learning task, 
Trigger continuously refines the model's fast hyper-parameters on the fly, boosting robustness under varying data conditions. 
We also design a Specialized Memory Updater ($\textit{Updater}$) that dynamically memorizes the context under the guidance of Trigger.
In experiments, Nirvana matches or surpasses strong LLM baselines on standard general-language benchmarks, while achieving the lowest perplexity across specialized domains including biomedicine, finance, and law.
On the challenging task of Magnetic Resonance Imaging (MRI), we attach lightweight codecs to the frozen Nirvana backbone and fine-tune them on paired k-space signals and images, achieving higher-fidelity reconstructions than conventional LLM-based models. In this setting, Trigger enables Nirvana to calibrate itself to the distribution of k-space signals and MRI images, yielding diagnostic-quality reconstructions and accurate clinical reports.
This unified approach of Trigger and Updater obviates the need for extensive domain-specific model backbone retraining, saving valuable time and data resources.
By seamlessly fusing broad linguistic intelligence with rapid, on-the-fly specialization, Nirvana ushers in a new class of general-to-special SGMs.

\begin{figure*}[t]
\centering
\centerline{\includegraphics[width=11.5cm]{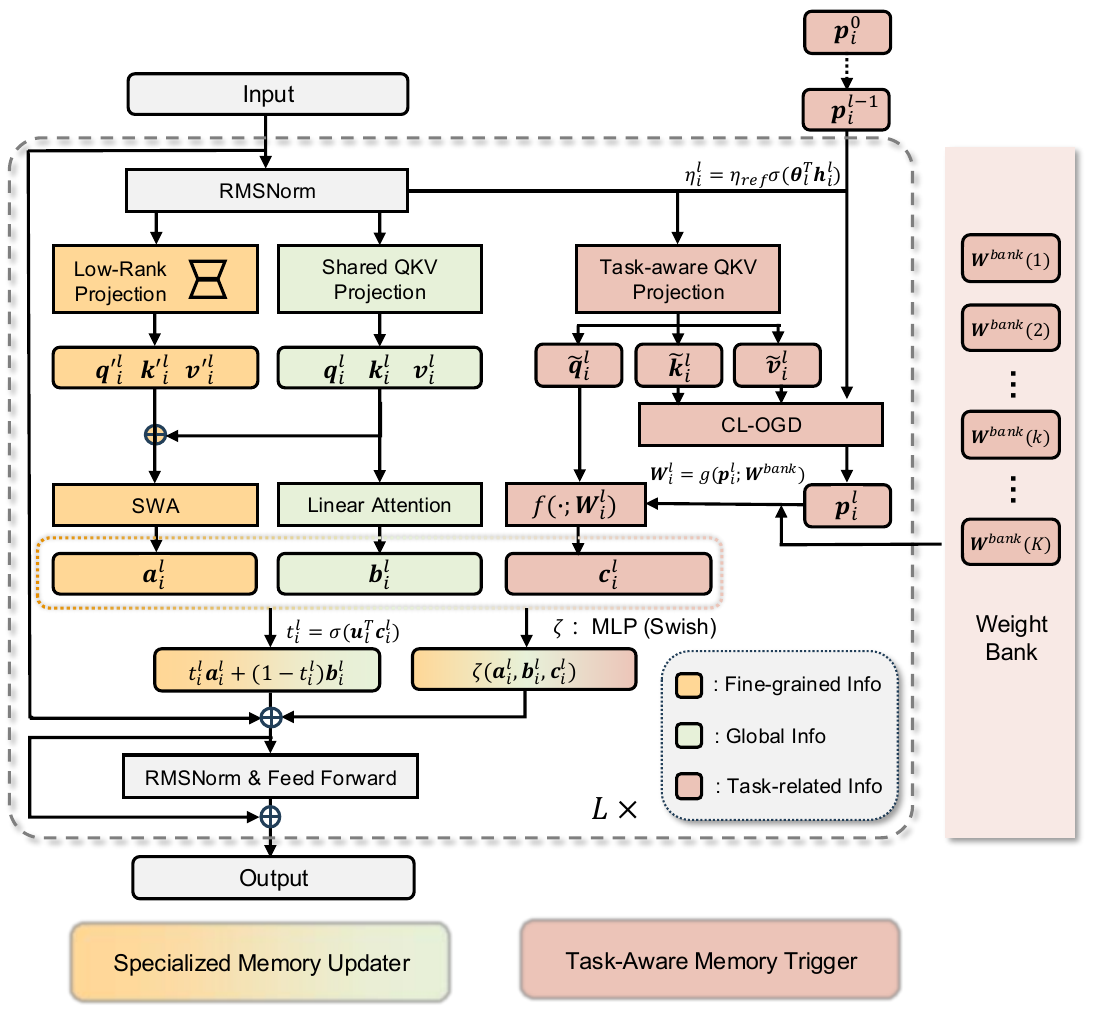}} 
  \caption{ Visualization of Nirvana's architecture. 
Updater employs conditional interpolation between SWA and linear attention, which can use any architecture from the 3-rd to the last but not least line in Table \ref{summary}.
Trigger extracts fast hyper-parameters $\boldsymbol{p}_i^l$ to update $f(\cdot ; \boldsymbol{W}_i^l)$ and generates task-related information $\boldsymbol{c}_i^l$ as the condition of Updater. 
  } 
  \label{nirvana}
\end{figure*}

\section{Related Work}

\paragraph{Hybrid Attention–Recurrent Architectures}
Recent work explores hybrid designs combining attention with recurrent architectures for long-context modeling. 
Samba \citep{samba} alternates Mamba with sliding-window attention to enable efficient sequence processing, though its fixed structure may limit adaptability.
Jamba \citep{jamba,jamba_1.5} introduces a Transformer--Mamba hybrid with MoE-style routing to achieve high throughput on long contexts, but relies on static gating and layer placement.
Gated DeltaNet \citep{yang2025gated} integrates gated recurrent updates into hybrid architectures (H1/H2), improving extrapolation and reasoning, but introduces additional inference overhead.

\paragraph{TTT and Meta-Learning}
Test-time training (TTT) \citep{sun2024learning} enables instance-wise adaptation via self-supervised updates during inference, improving robustness to distribution shift but incurring additional computation.
Meta-learning methods \citep{metalearning1,metalearning2} aim to acquire fast adaptation mechanisms but often struggle to generalize to streaming or per-sample settings.
Online meta-learning \citep{metalearning3} extends to continual learning scenarios, though it is not designed for efficient test-time self-supervision.
Building on these directions, Nirvana introduces a Trigger module that frames each test instance as an implicit self-supervised task, 
enabling lightweight adaptation under distribution shift with reduced overhead.

\section{Method}
To support specialized memory, we introduce a two-branch architecture operating at complementary levels.
The \textit{Trigger} branch captures abstract task signals, while the \textit{Updater} branch adaptively encodes task-specific context.
The two interact across layers, with Trigger conditioning Updater to interpolate between SWA and linear attention.
Figure~\ref{nirvana} presents the overall architecture of Nirvana.

\subsection{Specialized Memory Updater} 
SWA excels at modeling fine-grained information and local dependencies within a bounded context \citep{vaswani2017attention}, 
while linear attention enables accurate global information modeling of long sequences \citep{yang2025gated}.
Therefore, Updater is proposed to combine the advantages of both modules. 
We employ SWA instead of the full attention, such that the computational complexity of the model only grows \textit{linearly} with the length of the input sequence. 
With the aim of sharing the majority of parameters across the network, we use shared QKV projection matrices for SWA and linear attention. 
In order to learn the discrepancy of query, key, and value between the two modules with small computation overhead and few learnable parameters, we use a dimension reduction representation with low rank linear projection independently added before SWA. 
Denote ${\boldsymbol{q}'}_{i}^{l}$, ${\boldsymbol{k}'}_{i}^{l}$, and ${\boldsymbol{v}'}_{i}^{l}$ as the outputs of low rank linear projection, where the superscript $l$ and the subscript $i$ denote the $l$-th layer and the $i$-th token throughout this paper. 
We add ${\boldsymbol{q}'}_{i}^{l}$, ${\boldsymbol{k}'}_{i}^{l}$, and ${\boldsymbol{v}'}_{i}^{l}$ to the original counterparts to yield the query, key, and value of SWA.

\subsection{Task-Aware Memory Trigger}

The process of memory-related parametric learning can be viewed as compressing a massive training set into the weights of a model. 
This process of compression into the weights involves capturing the essence of the data that the model has been trained on. 
In conventional frameworks, the model's weights are shared across different tokens. 
However, Nirvana introduces a novel methodology that tailors the fast weights specifically for different layers and different tokens through Trigger. 



In order to facilitate the continuous flow of task-related information across various layers, we propose Trigger, which updates the fast weights by extracting the task in the context. 
Besides, we design a novel mechanism that allows the tokens not to share the same fast weights, such that the model can avoid information leakage during the training process. 
Specifically, the tokens have individual fast hyper-parameters \( \boldsymbol{p}_i^l \) that are implicitly determined by the task information.
In order to map \( \boldsymbol{p}_i^l \) to the fast weights of a neural network, we extract the tokens' individual fast weights from a fast weight bank \( \boldsymbol{W}^{\text{bank}} \), which is shared across different layers and different tokens. 
Generally, the fast weights are denoted by \( \boldsymbol{W}_i^l \in \mathbb{R}^{d\times d} \), which are generated from \( \boldsymbol{p}_i^l \) and \( \boldsymbol{W}^{\text{bank}} \) via a predefined function as:
\begin{align}
    \boldsymbol{W}_i^l = g(\boldsymbol{p}_i^l; \boldsymbol{W}^{\text{bank}}) .
    \label{gfunc}
\end{align}
Note that the generation of fast weights is conditional on both the layer and the token, allowing for a more granular control of the learning process across the network. 
Besides, the dimension of $\boldsymbol{p}_i^l$ should be much lower than $\boldsymbol{W}_i^l$, which ensures efficient parameter transfer across layers. 

To extract the abstract task-related information, we employ linear layers to compute the query, key, and value of Trigger, denoted by $\tilde{\boldsymbol{q}}_i^l$, $\tilde{\boldsymbol{k}}_i^l$, and $\tilde{\boldsymbol{v}}_i^l$, respectively. 
For computational efficiency, $\tilde{\boldsymbol{q}}_i^l$, $\tilde{\boldsymbol{k}}_i^l$, and $\tilde{\boldsymbol{v}}_i^l$ have a relatively low dimension compared to the dimension of hidden states. 
The task extraction process can be formulated as:
\begin{align}
    \boldsymbol{c}_i^l = f( \tilde{\boldsymbol{q}}_i^l ; \boldsymbol{W}_i^l) , 
\end{align}
where $f(\tilde{\boldsymbol{q}}_i^l ; \boldsymbol{W}_i^l)$ is a meta function modeled by a neural network that takes $\tilde{\boldsymbol{q}}_i^l$ as input and uses $\boldsymbol{W}_i^l$ as the network's test-time-changeable parameters. 
In order to update the fast hyper-parameters $\boldsymbol{p}_i^l$, we propose the Cross-Layer Online Gradient Descent (CL-OGD) algorithm.
CL-OGD guides Nirvana to update $f( \cdot ; \boldsymbol{W}_i^l)$ at the test time by minimizing the following loss function:
\begin{align}
    \mathcal{L}_i^l =\| f( \tilde{\boldsymbol{k}}_i^l ; \boldsymbol{W}_i^l) - \tilde{\boldsymbol{v}}_i^l \|^2_2 .
\end{align}

\begin{table*}[t]
\centering
\begin{tabular}{l c l c c}
\hline
\textbf{Term} & \textbf{Symbol} & \textbf{Role} & \textbf{Dimensionality} & \textbf{Updated at Test Time?} \\
\hline
Fast hyper-parameters & $p_i^l$ & Low-dim task signal & $K$ & Yes (via CL-OGD) \\
Fast weights & $W_i^l$ & Meta-function weights & $\mathbb{R}^{d\times d}$ & Yes (Indirectly) \\
Weight bank & $W_{\text{bank}}$ & Shared basis weights & $K$ blocks & No \\
\hline
\end{tabular}
\caption{Task-related terms and symbols in Nirvana.}
\label{fast_params_vs_weights}
\end{table*}

\begin{table*}[t]
  \centering
  \resizebox{\textwidth}{!}{
\begin{tabular}{l | cc | ccccccccc} 
\hline
\multirow{2}{*}{Model} & Wiki.  & LMB. & LMB. & PIQA & Hella. & Wino. & ARC-e  & ARC-c  & SIQA & BoolQ  & Avg. \\
& ppl $\downarrow$  & ppl $\downarrow$  & acc $\uparrow$ & acc $\uparrow$ & acc\_n $\uparrow$ & acc $\uparrow$ & acc $\uparrow$ & acc\_n $\uparrow$ & acc $\uparrow$ & acc $\uparrow$ & $\uparrow$  \\
\hline
Transformer++  & 18.53 & 18.32 & 42.60 & 70.02 & 50.23 & 53.51 & 68.83 & 35.10 & 40.66 & 57.09 & 52.25 \\
RetNet         & 19.08 & 17.27 & 40.52 & 70.07 & 49.16 & 54.14 & 67.34 & 33.78 & 40.78 & 60.39 & 52.02 \\
HGRN2          & 19.10 & 17.69 & 39.54 & 70.45 & 49.53 & 52.80 & 69.40 & 35.32 & 40.63 & 56.66 & 51.79 \\
Mamba          & 17.92 & 15.06 & 43.98 & 71.32 & 52.91 & 52.95 & 69.52 & 35.40 & 37.76 & 61.13 & 53.12 \\
Mamba2         & 16.56 & 12.56 & 45.66 & 71.87 & 55.67 & 55.24 & ${\textbf{72.47}}$ & 37.88 & 40.20 & 60.13 & 54.89 \\
DeltaNet       & 17.71 & 16.88 & 42.46 & 70.72 & 50.93 & 53.35 & 68.47 & 35.66 & 40.22 & 55.29 & 52.14 \\ 
Gated DeltaNet & 16.42 & 12.17 & 46.65 & $72.25$ & $55.76$ & $57.45$ & $71.21$ & $38.39$ & $40.63$ & 60.24 & $55.32$ \\ 
Samba          & 16.13 & 13.29 & 44.94 & 70.94 & 53.42 & 55.56 & 68.81 & 36.17 & 39.96 & ${62.11}$ & 54.00 \\
Gated DeltaNet-H1         & 16.07 & 12.12 & ${47.73}$ & ${72.57}$ & ${56.53}$ & ${58.40}$ & ${71.75}$ & ${\textbf{40.10}}$ & ${41.40}$ & ${{\textbf{63.21}}}$ & ${{56.40}}$ \\
Gated DeltaNet-H2         & \textbf{15.91} & 12.55 & ${48.76}$ & 72.19 & ${{56.88}}$ & ${57.77}$ & ${71.33}$ & ${39.07}$ & ${{\textbf{41.91}}}$ & 61.55 & ${56.18}$ \\
\hline 
Nirvana-noTrigger         & 16.60 & 12.25 & $49.40$ & $73.12$ & $57.43$ & $59.27$ & $68.80$ & $37.84$ & $41.50$ & $54.68$ & $55.26$ \\
\textbf{Nirvana (Ours)}   & 16.05 & \textbf{11.56} & $\textbf{50.37}$ & $\textbf{73.67}$ & $\textbf{58.25}$ & $\textbf{59.48}$ & $69.92$ & $39.51$ & $41.62$ & $59.27$ & $\textbf{56.51}$ \\
\hline 
\end{tabular}}
  \caption{Language modeling and zero-shot common sense reasoning performance of 1.3B models.}
\label{arc}
\end{table*}

Since the parameters of $f( \cdot ; \boldsymbol{W}_i^l)$ are decided by $\boldsymbol{p}_i^l$ according to Equation~\ref{gfunc}, updating $f( \cdot ; \boldsymbol{W}_i^l)$ in CL-OGD is equivalent to updating the fast hyper-parameters $\boldsymbol{p}_i^l$ at the test time, which can be formulated as:
\begin{align}
\Delta \boldsymbol{p}_i^l &=  \frac {\partial \mathcal{L}_i^l } {\partial \boldsymbol{W}_i^l} \frac {\partial \boldsymbol{W}_i^l} {\partial \boldsymbol{p}_i^l } =  \frac {\partial \mathcal{L}_i^l } {\partial \boldsymbol{W}_i^l} \frac {\partial g(\boldsymbol{p}_i^l; \boldsymbol{W}^{\text{bank}})} {\partial \boldsymbol{p}_i^l } , \label{delta} \\ 
\boldsymbol{p}_i^l & = \boldsymbol{p}_{i}^{l-1} -\eta_i^l \Delta \boldsymbol{p}_i^l , 
\label{ogd}
\end{align}
where $\eta_i^l$ is the adaptive online learning rate, defined as $\eta_i^l = \eta_{\text{ref}} \, \sigma(\boldsymbol{\theta}_{l}^\top \boldsymbol{h}_i^l)$ with $\boldsymbol{\theta}_{l}$ denoting a learnable projection vector and $\boldsymbol{h}_i^l$ denoting the hidden state before QKV projection. The function $\sigma(\cdot)$ denotes the Sigmoid function, and $\eta_{\text{ref}}$ is a reference learning rate.
In Nirvana, task-relevant information in the hidden states is extracted only after several prelude layers. Accordingly, in the first post-prelude layer, $\boldsymbol{p}_i^0$ is initialized as an all-1 vector for each token.

Since the hidden states vary in magnitude in different layers, we use Layer Normalization (LN) in $f(\boldsymbol{x}; \boldsymbol{W})$ for better stability.
To realize the relatively low computational complexity, $f(\boldsymbol{x} ; \boldsymbol{W})$ contains a linear layer, an LN operation, and a residual connection, i.e., $f(\boldsymbol{x}; \boldsymbol{W})=\boldsymbol{x}+\mathrm{LN}\left(f_{\text {linear }}(\boldsymbol{x}; \boldsymbol{W})\right)=\boldsymbol{x}+\mathrm{LN}\left(\boldsymbol{W}_{\text {linear}}\boldsymbol{x}+\boldsymbol{b}_{\text {linear }} \right)$. 
For the simplicity of calculation and for the ease of back propagation, 
we design a weight sharing mechanism that allows for the reuse of the same weight across different layers and tokens. 
The weight sharing mechanism is implemented through a learnable weight bank \( \boldsymbol{W}^{\text{bank}} \) that stores the shared weight parameters across different layers and tokens. 
The function $g(\boldsymbol{p}_i^l; \boldsymbol{W}^{\text{bank}})$ is formulated as:
\begin{align}
g(\boldsymbol{p}_i^l; \boldsymbol{W}^{\text{bank}}) = \sum_{k=1}^K \boldsymbol{p}_i^l(k) \boldsymbol{W}^{\text{bank}}(k),
\end{align}
where $\boldsymbol{p}_i^l(k)$ denotes the $k$-th element of $\boldsymbol{p}_i^l$ and $\boldsymbol{W}^{\text{bank}}(k)$ denotes the $k$-th block of $\boldsymbol{W}^{\text{bank}}$, respectively. 
Compared to the hidden states $\boldsymbol{h}_i^l$, $\boldsymbol{p}_i^l$ has a much smaller dimension (e.g., $K=64$), enabling efficient parameter transfer across layers.
Besides, $\boldsymbol{p}_i^l$ is updated according to Equation~\ref{delta} with the gradient computed as:
\begin{align}
\frac{\partial g(\boldsymbol{p}_i^l; \boldsymbol{W}^{\text{bank}})} {\partial \boldsymbol{p}_i^l } &= [\operatorname{vec}\{\boldsymbol{W}^{\text{bank}}(1)\} , \nonumber \\
& \ldots, \operatorname{vec}\{\boldsymbol{W}^{\text{bank}}(K)\}].
\end{align} 

Since task information is inherently high-level and difficult to extract within the early layers of Nirvana, we designate the first $N_{\text{pre}}$ layers as prelude layers, which operate without the Trigger and are not involved in task-information extraction. These prelude layers use only linear attention, while SWA is introduced in the subsequent post-prelude layers. Because standard linear attention architectures (e.g., Gated DeltaNet, Mamba2) are already capable of capturing position-dependent structure in the input sequence \citep{yang2025gated}, there is no need to apply Rotary Positional Embedding (RoPE) \citep{rope} before SWA. Incorporating RoPE at this stage would introduce unnecessary computation and could weaken the model’s ability to extrapolate to context lengths beyond those seen during training. Additional experiments and comparisons regarding RoPE are provided in Appendix~\ref{appdix_rope}.
The task-related symbols in Nirvana are summarized in Table~\ref{fast_params_vs_weights}.


The outputs of the SWA and the linear attention module are integrated by a conditional interpolation mechanism. 
Let the output of the SWA module be denoted by $\boldsymbol{a}_i^l$ and the output of the linear attention module be denoted by $\boldsymbol{b}_i^l$. 
The conditional interpolation mechanism is defined as
\begin{align}
    v_i^l = t_i^l \boldsymbol{a}_i^l + (1-t_i^l) \boldsymbol{b}_i^l + \zeta(\boldsymbol{a}_i^l,\boldsymbol{b}_i^l,\boldsymbol{c}_i^l),
\end{align}
where $t_i^l\in(0,1)$ is a task-dependent scalar that controls the interpolation between the outputs of the two modules. 
Specifically, $t_i^l$ is computed as $t_i^l = \sigma(\boldsymbol{u}_l^\top \boldsymbol{c}_i^l)$, where $\boldsymbol{u}_l$ is a learnable projection column vector. 
Besides, $\zeta(\boldsymbol{a}_i^l,\boldsymbol{b}_i^l,\boldsymbol{c}_i^l)$ adds a non-linear supplement to the conditional interpolation.
Specifically, $\zeta(\boldsymbol{a}_i^l,\boldsymbol{b}_i^l,\boldsymbol{c}_i^l)$ is a two-layer MLP with Swish activation function, and maps the concatenation of $\boldsymbol{a}_i^l,\boldsymbol{b}_i^l,\boldsymbol{c}_i^l$ to a vector at the same length of $\boldsymbol{a}_i^l$.
In order to make the number of the parameters in $\zeta(\boldsymbol{a}_i^l,\boldsymbol{b}_i^l,\boldsymbol{c}_i^l)$ relatively small, the length of the hidden layer in $\zeta(\boldsymbol{a}_i^l,\boldsymbol{b}_i^l,\boldsymbol{c}_i^l)$ is $1/8$ of the length of $\boldsymbol{a}_i^l$. 
The output of the conditional interpolation module,  $v_i^l$, is then passed into the subsequent RMSNorm and Feed-Forward Network (FFN).

\section{Experiments}

\begin{table*}[t]
  \centering
  \resizebox{14cm}{!}{
  \begin{tabular}{l | ccc ccc ccc c}
    \toprule
    \multirow{2}{*}{Model}
      & \multicolumn{3}{c}{S-NIAH-PK}
      & \multicolumn{3}{c}{S-NIAH-N}
      & \multicolumn{3}{c}{S-NIAH-W}
      & \multirow{2}{*}{Avg.} \\
    \cmidrule(lr){2-4} \cmidrule(lr){5-7} \cmidrule(lr){8-10}
      & 2K   & 4K   & 8K
      & 2K   & 4K   & 8K
      & 1K   & 2K   & 4K
      &       \\
    \midrule
    Transformer++  & $\textbf{100.0}$ & $\textbf{100.0}$  & 62.6 & $\textbf{100.0}$ & $\textbf{100.0}$ & 59.4 & $\textbf{100.0}$ & $\textbf{100.0}$ & \textbf{98.6} & 91.2 \\
    Mamba2         & 98.6 & 61.4 & 31.0 & 98.4 & 55.8 & 14.2 & 62.2 & 42.2 &  4.2 & 52.0 \\
    DeltaNet       & 96.8 & 98.8 & 98.6 & 47.2 & 15.4 & 12.8 & 85.2 & 46.2 & 20.0 & 57.9 \\
    Gated DeltaNet & 89.8 & 91.4 & 90.0 & 99.2 & 91.8 & 26.4 & 86.4 & 82.6 & 24.4 & 75.8 \\
    TTT            & 98.4 & 98.8 & 98.0 & 60.2 & 36.6 & 10.2 & 85.8 & 78.8 & 28.0 & 66.1 \\
    Samba          & 98.8 & 98.0 & 97.4 &  98.8 & 98.6 & 96.2 & 97.4 & 96.8 & 90.0 & 96.9  \\
    Gated DeltaNet-H2 & 99.2 & 97.8 & 97.4 & 98.0 & 97.8 & 96.2 & 98.0 & 97.4 & 96.8 & 97.6 \\
    \midrule
    Nirvana-noTrigger  & 99.6 & 99.6 & 99.0 & 99.8 & 99.8 & 98.8 & 99.0 & 97.4 & 94.8 & 98.6  \\
    \textbf{Nirvana (Ours)}        & \textbf{100.0} & $\textbf{100.0}$ & $\textbf{100.0}$ & $\textbf{100.0}$ & $\textbf{100.0}$ & $\textbf{99.6}$ & $98.8$ & $97.8$ & $95.4$ & $\textbf{99.1}$  \\
    \bottomrule
  \end{tabular}
  }
     \caption{S-NIAH performance of 1.3B models. S-NIAH-PK, S-NIAH-N, and S-NIAH-W are 3 tasks for single pass-key retrieval in a haystack, single number in a haystack, and single word in a haystack, respectively. All models are trained with 4K context length. }
    \label{niah}
\end{table*}

In experiments, we employ Gated DeltaNet \citep{yang2025gated} in the linear attention part of Nirvana, due to the outstanding performance of Gated DeltaNet in language modeling tasks. 
We train Nirvana from scratch with a training context window of length 4096 and a global batch size of 0.5M tokens. 
We use a model size of 1.3B parameters and train the model on 100B tokens sampled from the FineWeb dataset \citep{fineweb}. 
The window length of SWA in Updater is set as 2048. 
We employ the AdamW optimizer \citep{adamw} and a hybrid learning rate schedule of linear warm-up (the first 1B tokens) followed by the cosine decay, reaching a peak learning rate of $4\times 10^{-4}$.  
We utilize the LLaMA-2 tokenizer with a vocabulary size of 32,000. 
Training is conducted on 64 NVIDIA A800 GPUs.  
In evaluation, perplexity (ppl), accuracy (acc), and normalized accuracy (acc\_n) are measured with held-out test data on 8 NVIDIA A800 GPUs.
We also conduct the ablation study, where Nirvana-noTrigger refers to the Nirvana model without Trigger extracting the task-related information. 


\subsection{General Language Modeling}   

\subsubsection{Performance Comparison}   

In Table \ref{arc}, we report the models' language modeling performance using ppl on 2 datasets: Wikitext (Wiki.) and LAMBADA (LMB.),
and we also evaluate the models' zero-shot common sense reasoning performance using acc and acc\_n on 8 datasets: LAMBADA (LMB.), PIQA, HellaSwag (Hella.), WinoGrande (Wino.), ARC-easy (ARC-e), ARC-challenge (ARC-c), SIQA, and BoolQ. 
On Wiki. dataset, Nirvana achieves a ppl of 16.05, which is slightly higher than the SOTA model (15.91, Gated DeltaNet-H2). 
Notably, on LMB. dataset, Nirvana achieves a ppl of 11.56, which is better than the SOTA model (12.12, Gated DeltaNet-H1). 
On common sense reasoning tasks, Nirvana outperforms all the other models and achieves the highest accuracy on LMB., PIQA, Hella., and Wino. datasets. 
The performances of Nirvana on ARC-e, ARC-c, SIQA, and BoolQ are slightly worse than the SOTA models, but are still comparable. 
Moreover, Nirvana achieves the highest average accuracy on common sense reasoning tasks. 
In ablation study, the performance of Nirvana is better than Nirvana-noTrigger in terms of the average accuracy. 

We evaluate Nirvana on Single Needle-In-A-Haystack
(S-NIAH) benchmark with different context lengths according to RULER \citep{ruler}. 
In Table \ref{niah}, Nirvana outperforms all the other models in S-NIAH-PK and S-NIAH-N, and achieves the highest average accuracy. 
Particularly, in S-NIAH-PK (2K, 4K, and 8K context length) and in S-NIAH-N (2K and 4K context length), Nirvana achieves 100\% accuracy, remarkably higher than most of the existing models. 
When trained with 4K context length and tested with 8K context length,
Transformer++ does not perform well due to its relatively poor extrapolation ability.
However, Nirvana maintains its superior performance with 8K context length, which illustrates its solid extrapolation capability.
Notably, Nirvana-noTrigger performs worse than Nirvana, but is still better than other models on the average accuracy.

\color{black}
\subsubsection{Inference Efficiency} 

To quantify the computational complexity, we report the inference speed of all 1.3B models at a prompt length of 4096 tokens in  Table~\ref{tab:inference_speed}. 
Among all non-ablated models, Nirvana achieves the highest inference speed at 516 tokens/s, outperforming Samba (497 tokens/s), Gated DeltaNet (461 tokens/s), and Mamba2 (413 tokens/s), while substantially exceeding the classical Transformer++ baseline (191 tokens/s).
Nirvana-noTrigger achieves the highest inference speed at 568 tokens/s, but its slight advantage over full Nirvana comes with a substantial performance drop consistently observed across the earlier and later ablation experiments, clearly demonstrating the essential role of Trigger. 
Since Trigger operates on a compact 64-dimensional space, it adds only a negligible linear-time cost. 
Moreover, Updater’s SWA and linear attention modules maintain linear complexity, yielding faster inference than both Mamba2 and full-attention baselines. 
When combined with the accuracy gains in other experiments, these findings indicate that Nirvana improves model performance while simultaneously offering the best inference efficiency among the evaluated baselines. 

\begin{table}[t]
    \centering
            \resizebox{\columnwidth}{!}{%
    \begin{tabular}{l | c}
        \toprule
        Model & Inference Speed (tokens/s) \\
        \midrule
        Transformer++      & 191 \\
        Mamba2             & 413 \\
        Gated DeltaNet     & 461 \\
        Samba              & 497 \\
        \midrule
        Nirvana-noTrigger  & \textbf{568} \\
        Nirvana            & 516 \\
        \bottomrule
    \end{tabular}
    }
    \caption{\textcolor{black}{Inference speed of 1.3B models at prompt sequence length 4096, with batchsize = 4.}}
    \label{tab:inference_speed}
\end{table}

\color{black}

\subsection{Specialized Ability Evaluation}

\subsubsection{Performance in Specialized Domains: Biomedicine, Finance, and Law}

In order to assess performance in specialized domains, we evaluate various 1.3B models, including the ablated Nirvana-noTrigger, on three specialized corpora: (1) biomedical text from MIMIC-III clinical notes \citep{special1}, (2) financial news from April 2024 to October 2024 utilized in FinGPT \citep{special2}, and (3) legal documents from the Asylex refugee-status corpus \citep{special3}. All models are fine-tuned for 3 epochs on each domain. As shown in Table~\ref{tab:spec_domains}, {Nirvana} achieves the lowest perplexity in every domain and the best overall average of 7.78, substantially outperforming {Transformer++}, {Mamba2}, {Gated DeltaNet}, and Samba, whose averages range from 9.17 to 9.60. Importantly, the ablated {Nirvana-noTrigger} performs similarly to strong baselines but remains noticeably weaker than full {Nirvana} across all three domains, with an average perplexity gap of over 1.5 points. This consistent discrepancy highlights the essential role of Trigger in enabling {Nirvana} to adapt effectively to specialized-domain distributions, demonstrating that Trigger materially enhances domain-specific modeling beyond what the backbone alone can achieve. 

\begin{table}[t]
    \centering
        \resizebox{\columnwidth}{!}{%
    \begin{tabular}{l| cccc }
        \toprule
        \textbf{Model} & \textbf{Biomedicine} & \textbf{Finance} & \textbf{Law} & \textbf{Avg.} \\
        \midrule
        Transformer++      & 9.28 & 10.70 & 8.82 & 9.60 \\
        Mamba2             & 9.13 & 9.97  & 9.07 & 9.39 \\
        Gated DeltaNet     & 9.02 & 9.72  & 8.89 & 9.21 \\
        Samba              & 9.27 & 9.50  & 8.74 & 9.17 \\
        \midrule
        Nirvana-noTrigger  & 9.19 & 9.87  & 8.84 & 9.30 \\
        \textbf{Nirvana}   & \textbf{8.25} & \textbf{7.88} & \textbf{7.22} & \textbf{7.78} \\
        \bottomrule
    \end{tabular}
    }
        \caption{\textcolor{black}{Perplexity of 1.3B models on three specialized domains. Lower is better.}}
    \label{tab:spec_domains}
\end{table}

\begin{figure*}[t]
  \centering
  \begin{subfigure}[t]{0.32\linewidth}
    \centering
    \includegraphics[width=\linewidth]{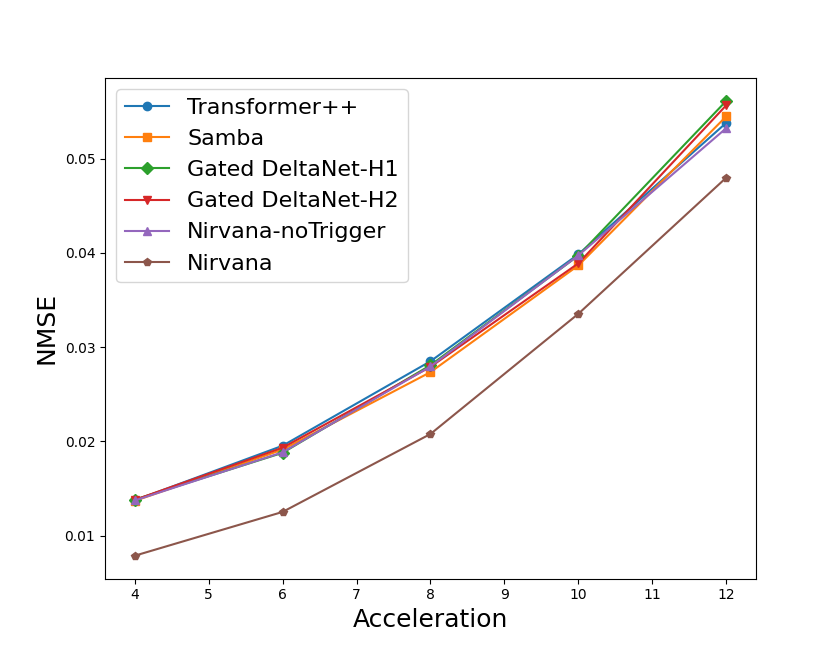}
    \subcaption{NMSE}
  \end{subfigure}
  \hfill
  \begin{subfigure}[t]{0.32\linewidth}
    \centering
    \includegraphics[width=\linewidth]{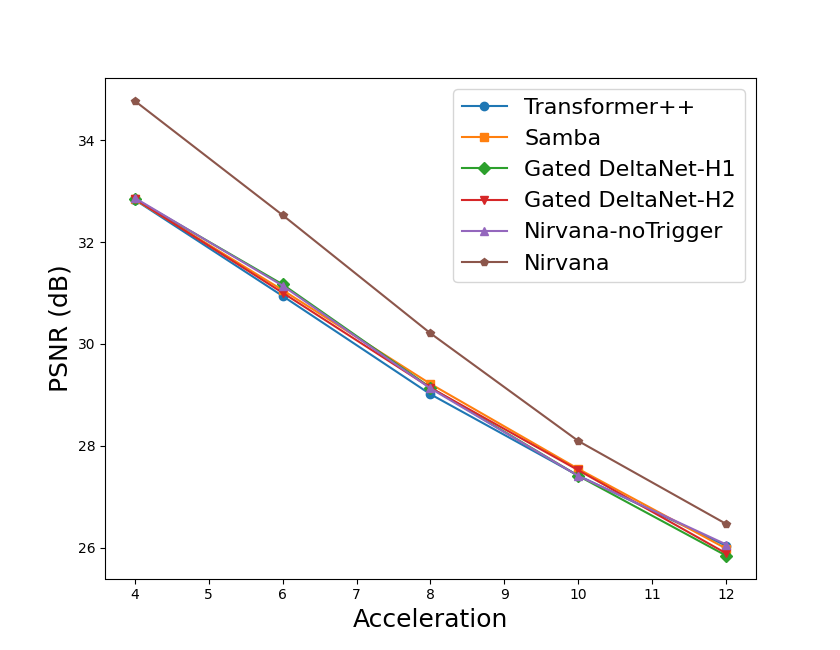}
    \subcaption{PSNR}
  \end{subfigure}
  \hfill
  \begin{subfigure}[t]{0.32\linewidth}
    \centering 
    \includegraphics[width=\linewidth]{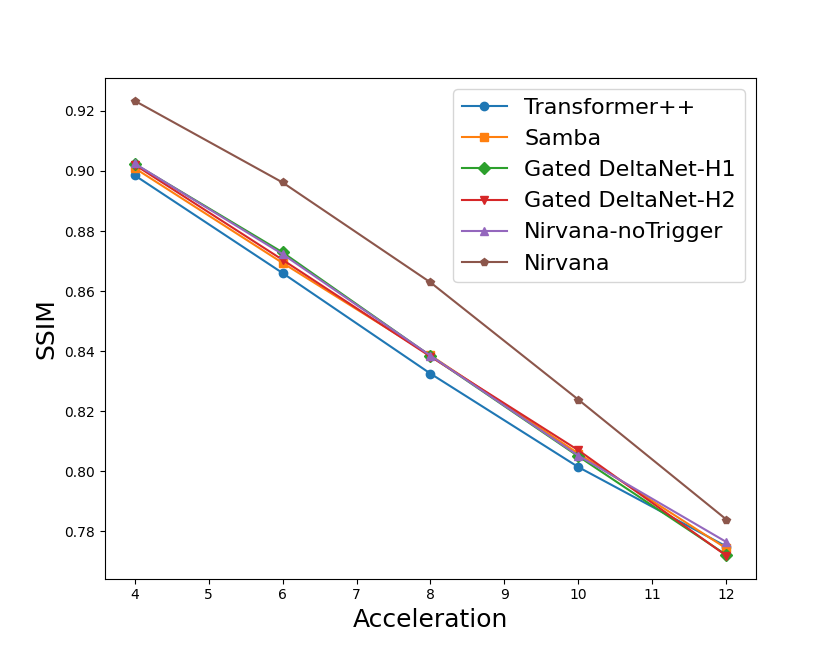}
    \subcaption{SSIM}
  \end{subfigure}
  \caption{\textcolor{black}{MRI reconstruction performance comparison between Nirvana and conventional LLMs with 160M trainable parameters in the k-space encoder and the image decoder. The acceleration rate is equivalent to the undersampling rate.}}
  \label{mri_plot_ablation}
\end{figure*}

\subsubsection{MRI Reconstruction and Report Generation}
\label{specialized}
MRI reconstruction is a clinically important yet technically challenging task that seeks to improve image quality while reducing scan time, making it a rigorous benchmark for evaluating Nirvana’s specialized capabilities. In this setting, Nirvana receives raw multi-coil k-space measurements along with an instruction prompt, and outputs both reconstructed image tokens and analysis tokens.
To map k-space data into Nirvana’s embedding space, we employ a multi-coil Variational Network (VarNet) \citep{varnet} followed by a lightweight ViT \citep{vit}, collectively serving as the k-space encoder. The encoder extracts k-space features and generates k-space tokens, which are concatenated with the instruction prompt and processed by Nirvana to generate image tokens and analysis tokens. A U-Net–based image decoder then converts the image tokens into the final reconstructed MRI images.
To mitigate instability caused by the limited amount of k-space data, we apply layer normalization before both the k-space encoder and the image decoder.

During post-training for MRI reconstruction, the language backbone of the 1.3B Nirvana model remains frozen, while training is applied solely to the k-space encoder and image decoder. 
The post-training process comprises two sequential stages. 
In the first stage, we only train the k-space encoder guided by the cross-entropy loss of only the generated MRI analysis tokens. 
After the k-space encoder is trained until convergence, we freeze the k-space encoder as well as the Nirvana backbone, such that the training in the second stage does not influence Nirvana's performance on MRI analysis. 
We then only train the image decoder with the MRI image reconstruction loss. 
Following \citep{varnet,udno}, the model minimizes the Structural Similarity Index Measure (SSIM) \citep{ssim} loss between the reconstructed image $\hat{\mathbf{x}}$ and the ground truth image $\mathbf{x}^*$ in the second stage:
\begin{equation}
\mathcal{L}_{2}\left(\hat{\mathbf{x}}, \mathbf{x}^*\right)=-\operatorname{SSIM}\left(\hat{\mathbf{x}}, \mathbf{x}^*\right). 
\end{equation}

To evaluate the performance of Nirvana for MRI reconstruction, we use the FastMRI dataset \citep{fastmri}. 
FastMRI dataset provides paired k-space signals and MRI images that can be directly used in the second post-training stage.
In the first post-training stage, we create a list of possible instruction prompts, such as ``According to the k-space signals, are there any pathological features?"
The ground truth analysis of the MRI images corresponding to the instruction prompt is generated by the Lingshu Model \citep{lingshu}. 

MRI reconstruction is greatly limited by a slow data acquisition process, which sometimes requires patients to remain still for an hour \citep{chen2022ai,singh2023emerging}.  
Thus, it is essential to accelerate the MRI scan by undersampling in the scanning process. 
Following \citep{fastmri,varnet,udno}, we undersample the k-space signals in the frequency domain to accelerate MRI acquisition while reducing the amount of data to be processed. Detailed configurations are shown in Appendix~\ref{appdix_mri}.

\color{black}
We compare the MRI reconstruction performance of 1.3B Nirvana with other 1.3B LLMs across different undersampling rates in Figure~\ref{mri_plot_ablation}. All models are pretrained on FineWeb and post-trained with the frozen backbone using the same procedure described at the beginning of Section~\ref{specialized}. Nirvana consistently achieves higher reconstruction fidelity across all settings. In the ablation study, Nirvana-noTrigger shows substantially degraded performance, highlighting the critical role of Trigger in enabling effective adaptation for MRI reconstruction.
\color{black}


\begin{figure}[t]
  \centering
\centerline{\includegraphics[width=7.0cm]{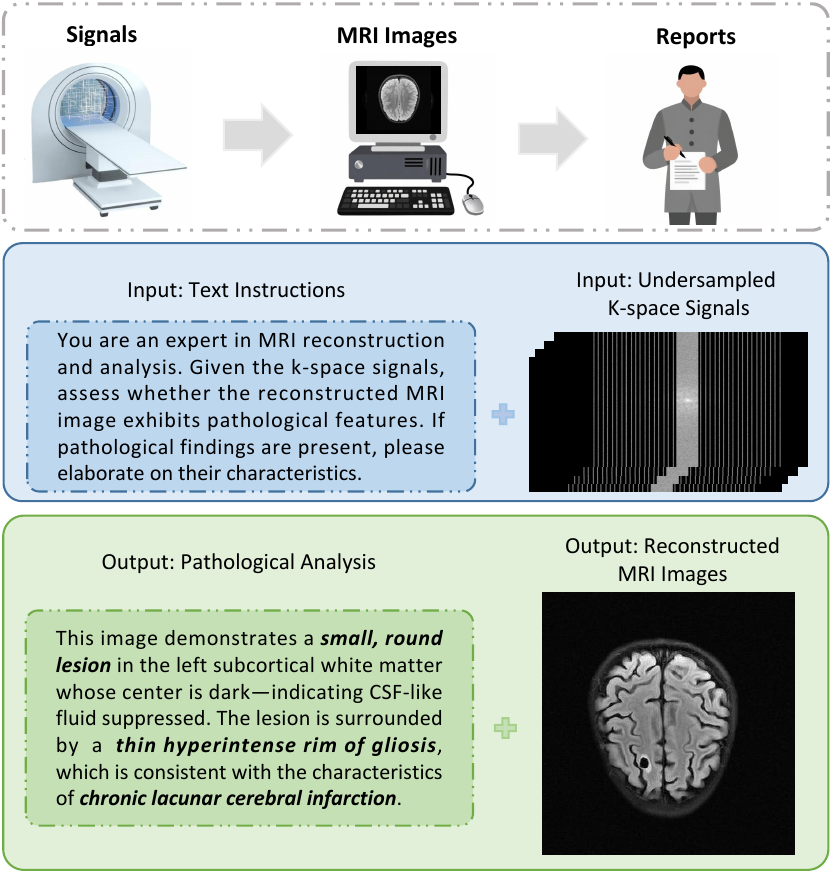}}   
  \caption{\textcolor{black}{The overall process of MRI reconstruction and report generation by Nirvana.}}
  \label{report}
\end{figure}

We show an example of the overall MRI reconstruction and report generation by Nirvana in Figure~\ref{report}. 
Nirvana takes the undersampled k-space signals and the instruction prompt as input, and outputs the reconstructed MRI image as well as the corresponding analysis. 
In contrast with traditional MRI report generation models (such as Lingshu \citep{lingshu}, HealthGPT \citep{healthgpt}, and MedGemma \citep{medgemma}) which directly take the reconstructed MRI image as input, Nirvana takes the k-space signals undersampled from the raw signals received by the coils as input to generate the overall MRI report, including the reconstructed image. 
In Figure~\ref{report}, the report generated by Nirvana accurately captures the important pathological features of the image, including the lesion's shape, size, position, and surrounding matter. 
Moreover, the report provides further diagnosis that the reconstructed MRI image is consistent with the characteristics of chronic lacunar cerebral infarction. 
More experiment results and analysis of MRI are shown in Appendix~\ref{appdix_mri}.

\subsection{Fast Parameter Evolution Analysis}
\label{sec:fast_param_evolution}

To better understand the behavior of Trigger, we analyze how the fast hyper-parameters $p_i^l \in \mathbb{R}^{K}$ evolve during inference.
We log $p_i^l$ and $\eta_i^l$ during inference on a held-out mixture of general-domain and specialized-domain inputs, including FineWeb validation slices and the biomedical, finance, and law corpora used in Section~4.2.1.
Statistics are aggregated over 1,024 sequences of length 4096 using greedy decoding with batch size 4.
We report the layer-wise relative update magnitude
\begin{equation}
r_l = \mathbb{E}_i \left[\frac{\lVert p_i^l - p_i^{l-1}\rVert_2}{\lVert p_i^{l-1}\rVert_2}\right],
\end{equation}
the layer-wise cosine change
\begin{equation}
d_l = \mathbb{E}_i\left[1 - \cos\left(p_i^l, p_i^{l-1}\right)\right],
\end{equation}
and the token-wise average update speed
\begin{equation}
s(i)=\frac{1}{L-N_{\text{pre}}}\sum_{l=N_{\text{pre}}+1}^{L}\frac{\lVert p_i^l - p_i^{l-1}\rVert_2}{\lVert p_i^{l-1}\rVert_2}.
\end{equation}
We also report the mean and P95 of $\eta_i^l$, together with the large-update rate, i.e., the fraction of tokens with $s(i)>0.12$.

\paragraph{Layer-wise dynamics.}
Table~\ref{tab:layerwise_fast_param} summarizes the layer-wise statistics.
Two trends are clear.
First, fast adaptation is concentrated immediately after the prelude stage: the first post-prelude layer yields the largest update magnitude and directional change, consistent with our design that task-relevant signals become extractable only after several prelude layers.
Second, both $r_l$ and $d_l$ decay monotonically with depth, while $\eta_i^l$ follows the same trend.
This suggests that Trigger performs rapid early task alignment and then transitions to a stable corrective regime.
Importantly, even the upper-tail updates remain moderate in late layers, indicating bounded inference-time adaptation.

\begin{table}[t]
\centering
\resizebox{\columnwidth}{!}{%
\begin{tabular}{lcccccc}
\toprule
Layer bucket & Trigger & Mean $r_l$ & P95 $r_l$ & Mean $d_l$ & Mean $\eta_i^l$ & P95 $\eta_i^l$ \\
 & active? &  &  &  & ($\times 10^{-3}$) & ($\times 10^{-3}$) \\
\midrule
1--4 (Prelude) & $\times$ & 0.000 & 0.000 & 0.000 & 0.00 & 0.00 \\
5 (1st post-prelude) & $\checkmark$ & 0.182 & 0.356 & 0.041 & 1.78 & 3.42 \\
6--8 & $\checkmark$ & 0.112 & 0.231 & 0.026 & 1.41 & 2.87 \\
9--12 & $\checkmark$ & 0.076 & 0.162 & 0.018 & 1.18 & 2.45 \\
13--16 & $\checkmark$ & 0.058 & 0.129 & 0.014 & 1.02 & 2.18 \\
17--20 & $\checkmark$ & 0.043 & 0.103 & 0.010 & 0.91 & 1.95 \\
21--24 (Late) & $\checkmark$ & 0.034 & 0.086 & 0.008 & 0.84 & 1.82 \\
\bottomrule
\end{tabular}%
}
\caption{Layer-wise fast hyper-parameter dynamics.}
\label{tab:layerwise_fast_param}
\end{table}

\paragraph{Token-wise dynamics.}
To study positional effects, we partition the 4096-token context into contiguous 512-token buckets and compute token-wise update statistics.
Results are shown in Table~\ref{tab:tokenwise_fast_param}.
We observe a clear front-loaded adaptation pattern.
Early tokens exhibit the largest update speed, highest large-update rate, and largest adaptive learning rate, while all three quantities decrease smoothly with position.
This indicates that Trigger mainly performs task calibration near the beginning of the sequence, and then maintains a progressively more conservative regime over long contexts.
Notably, no secondary peaks appear in later regions, suggesting that fast hyper-parameter evolution remains stable rather than repeatedly re-entering a highly plastic state.

\begin{table}[t]
\centering
\resizebox{\columnwidth}{!}{%
\begin{tabular}{lcccc}
\toprule
Token position bucket & Mean $s(i)$ & P95 speed & Large-update rate & Mean $\eta$ over layers \\
 &  &  & $s(i)>0.12$ & ($\times 10^{-3}$) \\
\midrule
1--512 & 0.094 & 0.211 & 24.8\% & 1.23 \\
513--1024 & 0.071 & 0.168 & 16.2\% & 1.11 \\
1025--1536 & 0.059 & 0.143 & 12.4\% & 1.02 \\
1537--2048 & 0.052 & 0.129 & 9.6\% & 0.97 \\
2049--2560 & 0.048 & 0.118 & 8.1\% & 0.94 \\
2561--3072 & 0.046 & 0.112 & 7.5\% & 0.92 \\
3073--3584 & 0.046 & 0.110 & 6.9\% & 0.91 \\
3585--4096 & 0.043 & 0.105 & 6.5\% & 0.89 \\
\bottomrule
\end{tabular}%
}
\caption{Token-wise fast hyper-parameter dynamics .}
\label{tab:tokenwise_fast_param}
\end{table}

Overall, fast hyper-parameter dynamics are both depth-aware and position-aware: adaptation is strongest immediately after the prelude layers and near the beginning of the sequence, then gradually stabilizes across both depth and position.

\section{Conclusion}

In this work, we present Nirvana, an SGM with the task-aware memory mechanism. By enabling dynamic interpolation between SWA and linear attention, Updater allows the model to flexibly balance local and global information flow while maintaining computational efficiency.
Complementing this, Trigger introduces per-sample self-supervision, allowing Nirvana to adapt to distributional shifts without requiring backbone retraining.
Experiments show that Nirvana matches or exceeds strong LLM baselines on general benchmarks, and furthermore achieves the lowest perplexity across specialized domains including biomedicine, finance, and law. In the challenging MRI task, Nirvana yields higher-fidelity reconstructed MRI images than conventional LLM-based models, and it also generates reliable preliminary clinical reports. Importantly, ablation studies reveal that removing Trigger leads to notable performance degradation across all evaluation tasks, demonstrating its essential role in task-aware specialization. These findings indicate that Nirvana can transition smoothly from general language understanding to diverse specialized and high-precision domains.

\section{Limitations}

We summarize the limitations of Nirvana as follows.
First, our current study is still restricted to a moderate-scale setting, namely a 1.3B-parameter model trained on 100B tokens.
While this scale is sufficient to validate the core design and to enable controlled comparisons across multiple variants, it remains substantially smaller than the scale of frontier foundation models.
As a result, our findings should be interpreted primarily as evidence of the effectiveness of the proposed architecture at this scale, rather than as a definitive characterization of its scaling behavior.
Extending Nirvana to substantially larger model sizes or token budgets may require further system-level and engineering optimizations.
That said, we believe the core design is largely orthogonal to such optimizations and can benefit from future advances in efficient training and inference.

Second, we have not exhaustively explored all architectural variants or hyperparameter configurations.
Although our design choices follow standard and principled practices, more systematic tuning may further improve performance, stability, and efficiency under different model scales and task regimes.

Third, the current MRI evaluation remains limited in scope.
In particular, our experiments rely on synthetic undersampling settings and automatically generated reports, which do not fully capture the complexity of real-world clinical workflows.
Therefore, while the MRI results provide an initial indication that Nirvana can extend beyond standard language modeling benchmarks, they should not be interpreted as a comprehensive clinical validation.
A more complete assessment would require evaluation under real acquisition pipelines, broader medical datasets, and expert-verified reporting settings.

Despite these limitations, we believe that our work provides a solid foundation for studying test-time adaptive memory in language models and beyond.
Future work will explore larger-scale settings, broader architectural and hyperparameter spaces, and more realistic specialized-domain evaluations to further assess the capability and robustness of Nirvana.

\section{Ethical Considerations}

Our research has been conducted with a clear commitment to avoiding harm and upholding honesty and transparency in both methodology and reporting. We have made deliberate efforts to identify and mitigate potential biases in data and algorithms to promote fairness. In addition, we have respected individual privacy and ensured full compliance with all relevant regulations and ethical standards governing data use. 

\section*{Acknowledgments}
This work was supported by the National Natural Science Foundation of China (Grant No. 6250076080, 62325107 and U23A20272), and supported by the China Postdoctoral Science Foundation under Grant Number 2025M771537, and supported by the Science and Technology Department of Sichuan Province (Grant 2025YFHZ0022).

\bibliography{custom}

\newpage

\appendix

\section*{Appendices}

\section{Reproducibility Statement}

In this paper, we present a novel SGM called Nirvana.
To guarantee that our work can be easily reproduced and built upon by the research community, we have taken the following key steps.
First, the source code implementing our method 
is available as part of the supplementary materials.
The code includes all scripts necessary for training and evaluating Nirvana, while pretrained models will be released after the reviewing process.
Experimental settings and hyperparameters are available in our experiments and appendices.
We use publicly available datasets for training and evaluation, and details are reported in the experiments. 
Finally, we also provide information about the hardware environment used in our experiments.
Models are available at \url{https://huggingface.co/collections/YuhuaJiang/nirvana}.
Code is available at \url{https://github.com/YuhuaJiang2002/Nirvana}.


\appendix

\section{Statement for Use of LLMs}

LLMs were only used to assist with language polishing in certain sections of this paper.

\section{Detailed Configurations}
\label{app:config}

In this appendix, we provide the detailed configurations of the full 1.3B Nirvana model, the CL-OGD adaptation module, and the MRI setting used in our experiments. These details are included to improve reproducibility and to clarify the architectural and optimization choices underlying the reported results.

\subsection{Full Model Configuration}
\label{app:full_model_config}

Table~\ref{tab:nirvana_full_config} summarizes the full configuration of the 1.3B Nirvana model. The model is built with 24 layers and a hidden size of 2048, resulting in a total parameter count of 1.30B. We use a context length of 4096, with SWA applied over a local window of size 2048, while long-range dependencies are handled by the linear attention component instantiated with Gated DeltaNet. The fast hyper-parameter dimension and the number of weight bank blocks are both set to 64. For optimization, we adopt AdamW with a peak learning rate of $4 \times 10^{-4}$, trained on 100B tokens with a 1B-token warmup and a global batch size of 0.5M tokens over 64 A800 GPUs.

\begin{table}[t]
\centering
\begin{tabular}{ll}
\toprule
\textbf{Component} & \textbf{Value} \\
\midrule
Total Parameters & 1.30B \\
Hidden Size $d$ & 2048 \\
FFN Dimension & 5632 \\
Number of Layers $L$ & 24 \\
Prelude Layers $N_{\text{pre}}$ & 4 \\
Attention Heads & 16 \\
Head Dimension & 128 \\
Context Length & 4096 \\
SWA Window Size & 2048 \\
Linear Attention Mechanism & Gated DeltaNet \\
Fast Hyper-Parameter Dim $K$ & 64 \\
Weight Bank Blocks & 64 \\
Trigger MLP Hidden Ratio & $1/8$ of $d$ \\
Interpolation MLP Activation & Swish \\
Positional Encoding & No RoPE in SWA \\
Optimizer & AdamW \\
Peak LR & $4 \times 10^{-4}$ \\
Warmup & 1B tokens \\
Training Tokens & 100B \\
Global Batch Size & 0.5M tokens \\
GPUs & 64 $\times$ A800 \\
\bottomrule
\end{tabular}
\caption{Full configuration of the 1.3B Nirvana model.}
\label{tab:nirvana_full_config}
\end{table}

\subsection{CL-OGD Configuration}
\label{app:clogd_config}

Table~\ref{tab:clogd_config} reports the configuration of CL-OGD used in our framework. The reference learning rate is set to $1 \times 10^{-3}$, while the actual update step is modulated adaptively through a sigmoid gate, i.e., $\eta_i^l = \eta_{\text{ref}} \sigma(\theta_l^\top h_i^l)$. This design allows the update magnitude to vary across layers and tokens according to the inferred task signal. The trigger module uses a QKV dimension of 512, and the online update objective is defined as a squared $\ell_2$ regression loss between the transformed key representation and the target value representation.

\begin{table}[t]
\centering
\begin{tabular}{ll}
\toprule
\textbf{Parameter} & \textbf{Value} \\
\midrule
Reference LR $\eta_{\text{ref}}$ & $1 \times 10^{-3}$ \\
Adaptive LR & $\eta_i^l = \eta_{\text{ref}} \sigma(\theta_l^\top h_i^l)$ \\
Trigger QKV Dim & 512 \\
Update Loss & $L_i^l = \| f(\tilde{k}_i^l; W_i^l) - \tilde{v}_i^l \|_2^2$ \\
\bottomrule
\end{tabular}
\caption{Configuration of CL-OGD used in Nirvana.}
\label{tab:clogd_config}
\end{table}

\subsection{MRI Configuration}
\label{app:mri_config}

For the MRI experiments, we adopt a frozen Nirvana 1.3B backbone and introduce 160M trainable parameters for task-specific adaptation. As shown in Table~\ref{tab:mri_config}, the encoder consists of VarNet and ViT components, while the decoder is implemented with a U-Net architecture. The reconstruction objective is defined using negative SSIM, which directly encourages structural fidelity in the reconstructed image. Experiments are conducted on the FastMRI dataset under a range of undersampling ratios from $4\times$ to $12\times$.

\begin{table}[t]
\centering
\begin{tabular}{ll}
\toprule
\textbf{Component} & \textbf{Setting} \\
\midrule
Backbone & Frozen Nirvana 1.3B \\
Trainable Params & 160M \\
Encoder & VarNet + ViT \\
Decoder & U-Net \\
Loss & $-\mathrm{SSIM}(\hat{x}, x^*)$ \\
Dataset & FastMRI \\
Undersampling & $4\times, 6\times, 8\times, 10\times, 12\times$ \\
\bottomrule
\end{tabular}
\caption{Configuration for the MRI experiments.}
\label{tab:mri_config}
\end{table}

\begin{table*}[t]
  \centering
  \begin{tabular}{l | ccc ccc ccc c}
    \toprule
    \multirow{2}{*}{Model}
      & \multicolumn{3}{c}{S-NIAH-PK}
      & \multicolumn{3}{c}{S-NIAH-N}
      & \multicolumn{3}{c}{S-NIAH-W}
      & \multirow{2}{*}{Avg.} \\
    \cmidrule(lr){2-4} \cmidrule(lr){5-7} \cmidrule(lr){8-10}
      & 2K   & 4K   & 8K
      & 2K   & 4K   & 8K
      & 1K   & 2K   & 4K
      &       \\
    \midrule
    Nirvana-RoPE   & $\textbf{100.0}$ & $\textbf{100.0}$ & $0.2$ & $\textbf{100.0}$ & $\textbf{100.0}$ & $4.4$ & $\textbf{100.0}$ & $97.0$ & $92.8$ & $77.2$ \\
    \textbf{Nirvana (Ours)}        & \textbf{100.0} & $\textbf{100.0}$ & $\textbf{100.0}$ & $\textbf{100.0}$ & $\textbf{100.0}$ & $\textbf{99.6}$ & $98.8$ & $\textbf{97.8}$ & $\textbf{95.4}$ & $\textbf{99.1}$  \\
    \bottomrule
  \end{tabular}
      \caption{S-NIAH performance of 1.3B Nirvana with or without RoPE. S-NIAH-PK, S-NIAH-N, and S-NIAH-W are 3 tasks for the pass-key retrieval in a haystack, number in a haystack, and word in a haystack, respectively. Both models are trained with 4K context length. }
    \label{niah2}
\end{table*}

\begin{table*}[t]
  \centering
  \resizebox{\textwidth}{!}{
\begin{tabular}{l | cc | ccccccccc} 
\hline
Model & Wiki.  & LMB. & LMB. & PIQA & Hella. & Wino. & ARC-e  & ARC-c  & SIQA & BoolQ  & Avg. \\
& ppl $\downarrow$  & ppl $\downarrow$  & acc $\uparrow$ & acc $\uparrow$ & acc\_n $\uparrow$ & acc $\uparrow$ & acc $\uparrow$ & acc\_n $\uparrow$ & acc $\uparrow$ & acc $\uparrow$ & $\uparrow$  \\
\hline
Nirvana-RoPE              & 18.01 & 12.13 & $49.97$ & $\textbf{73.71}$ & $58.17$ & $58.43$ & $68.90$ & $\textbf{38.86}$ & $41.15$ & $\textbf{59.33}$ & $56.07$ \\ 
\textbf{Nirvana (Ours)}   & \textbf{17.57} & \textbf{11.56} & $\textbf{50.37}$ & $73.67$ & $\textbf{58.25}$ & $\textbf{59.48}$ & $\textbf{68.92}$ & $38.51$ & $\textbf{41.62}$ & $59.27$ & $\textbf{56.26}$ \\
\hline 
\end{tabular}}
  \caption{Language Modeling and Zero-Shot Common Sense Reasoning Performance of 1.3B Nirvana with or without RoPE.}
\label{arc2}
\end{table*}

\subsection{CL-OGD Pseudocode}
\label{app:clogd_pseudocode}

For completeness, Algorithm~\ref{alg:clogd} summarizes the Cross-Layer Online Gradient Descent (CL-OGD) procedure used in the Trigger branch.
For clarity, we re-index the post-prelude layers as $l=1,\dots,L_{\text{post}}$, where $L_{\text{post}}=L-N_{\text{pre}}$, and reserve $p_i^0$ for the initialization before the first post-prelude layer.
The key idea is to update a low-dimensional fast hyper-parameter vector $p_i^l \in \mathbb{R}^{K}$ for each token, rather than directly optimizing the full fast weights in $\mathbb{R}^{d\times d}$.
At each post-prelude layer, the current hyper-parameter vector is first mapped to the fast weights through the shared weight bank $W^{\text{bank}}$, then refined by an online gradient step using the local key-value regression objective.
The updated fast hyper-parameters are subsequently used to produce the task signal $c_i^l$, which conditions the interpolation between SWA and linear attention in the Updater.

\begin{algorithm}[t]
\caption{CL-OGD in the Trigger branch}
\label{alg:clogd}
\small
\begin{algorithmic}[1]
\Require Post-prelude hidden states $\{h_i^l\}_{l=1,i=1}^{L_{\text{post}},T}$, reference learning rate $\eta_{\text{ref}}$, shared weight bank $W^{\text{bank}}$, Trigger projections $\{\Theta_Q^l,\Theta_K^l,\Theta_V^l,\theta_l\}_{l=1}^{L_{\text{post}}}$, meta-function $f(\cdot;W)$, bank mapping $g(\cdot;W^{\text{bank}})$
\Ensure Task signals $\{c_i^l\}$ and updated fast hyper-parameters $\{p_i^l\}$

\State Initialize $p_i^0 \leftarrow \mathbf{1}\in\mathbb{R}^{K}$ for all tokens $i=1,\dots,T$

\For{$l=1$ to $L_{\text{post}}$}
    \For{$i=1$ to $T$}
        \State $\tilde{q}_i^l \leftarrow \Theta_Q^l h_i^l$, \quad
               $\tilde{k}_i^l \leftarrow \Theta_K^l h_i^l$, \quad
               $\tilde{v}_i^l \leftarrow \Theta_V^l h_i^l$
        \State $W_{i,\mathrm{pre}}^l \leftarrow g(p_i^{\,l-1};W^{\text{bank}})
               = \sum_{k=1}^{K} p_i^{\,l-1}(k)\,W^{\text{bank}}(k)$
        \State $\mathcal{L}_i^l \leftarrow
               \left\| f(\tilde{k}_i^l;W_{i,\mathrm{pre}}^l)-\tilde{v}_i^l \right\|_2^2$
        \State $\eta_i^l \leftarrow \eta_{\text{ref}}\,\sigma\!\left((\theta_l)^\top h_i^l\right)$
        \State $\Delta p_i^l \leftarrow
               \dfrac{\partial \mathcal{L}_i^l}{\partial W_{i,\mathrm{pre}}^l}
               \dfrac{\partial g(p_i^{\,l-1};W^{\text{bank}})}{\partial p_i^{\,l-1}}$
        \State $p_i^l \leftarrow p_i^{\,l-1} - \eta_i^l \Delta p_i^l$
        \State $W_i^l \leftarrow g(p_i^l;W^{\text{bank}})$
        \State $c_i^l \leftarrow f(\tilde{q}_i^l;W_i^l)$
    \EndFor
\EndFor

\State \Return $\{c_i^l\}, \{p_i^l\}$
\end{algorithmic}
\end{algorithm}

Algorithm~\ref{alg:clogd} highlights that CL-OGD performs test-time adaptation in a lightweight manner.
Instead of updating the full fast weights directly, Nirvana only updates the compact hyper-parameters $p_i^l$, while the shared weight bank $W^{\text{bank}}$ remains fixed.
The fast weights are then generated on the fly as a linear combination of the bank blocks, which makes the adaptation both parameter-efficient and computationally tractable.
Moreover, the recursion $p_i^{l-1}\rightarrow p_i^l$ enables cross-layer propagation of task information, allowing the Trigger branch to progressively refine token-wise task signals as depth increases.
The resulting $c_i^l$ is finally used by the Updater to compute the interpolation coefficient $t_i^l=\sigma(u_l^\top c_i^l)$ and thus modulate the balance between SWA and linear attention in a task-dependent manner.

For brevity, Algorithm~\ref{alg:clogd} is written for a single sequence and an explicit token-wise loop.
In practice, all tokens within a layer are processed in parallel on modern accelerators.

\section{Contrast With RAG and External Memory}
\label{app:rag_comparison}

In this section, we clarify the relationship between Nirvana and retrieval-based or external-memory-based methods, especially Retrieval-Augmented Generation (RAG).
Although both paradigms can improve a model's ability to handle knowledge-intensive or domain-specific inputs, they operate through fundamentally different mechanisms.

RAG augments the current input by retrieving relevant documents from an external database and concatenating them to the context.
Its memory is therefore \emph{non-parametric}: the additional knowledge remains outside the model parameters and is accessed through explicit retrieval.
In contrast, Nirvana does not require an external index or database.
Instead, it performs \emph{parametric} test-time adaptation by updating the token-wise fast hyper-parameters $p_i^l$, which in turn determine the fast weights $W_i^l$ used by the Trigger:
\begin{equation}
p_i^l \rightarrow W_i^l \rightarrow f(\cdot; W_i^l).
\end{equation}
As a result, RAG changes the \emph{input context}, whereas Nirvana changes the model's \emph{internal processing pathway}.

\begin{table*}[t]
\centering
\begin{tabular}{lll}
\toprule
\textbf{Aspect} & \textbf{RAG} & \textbf{Nirvana} \\
\midrule
External database & Required & Not required \\
Memory type & Non-parametric & Parametric fast weights \\
Adaptation mechanism & Retrieval-based & Gradient-based \\
Latency & Retrieval overhead & Linear-time update \\
Knowledge update & Static index lookup & Online adaptation \\
\bottomrule
\end{tabular}
\caption{Comparison between RAG and Nirvana.}
\label{tab:rag_vs_nirvana}
\end{table*}

Table~\ref{tab:rag_vs_nirvana} summarizes the main differences.
This distinction is important conceptually.
RAG relies on external storage and explicit access to retrieved evidence, which can be highly effective when a reliable retrieval corpus is available.
Nirvana, by contrast, specializes its internal computation on the fly through CL-OGD, without introducing external retrieval latency or requiring index construction and maintenance.
Its adaptation is therefore more naturally viewed as a form of dynamic internal memory rather than external lookup.

At the same time, the two approaches are not competing alternatives, but largely complementary.
RAG can provide explicit external evidence, while Nirvana can adapt the model's internal processing to better absorb and utilize the current context.
In this sense, a retrieval module and Nirvana-style fast adaptation could in principle be combined, with retrieval enriching the input and Trigger/Updater refining the model's internal response to that input.

\section{Contrast With Miras}
\label{app:miras_comparison}

Miras \citep{behrouz2025s} offers a thoughtful and technically solid study of the connections among test-time memorization, attentional bias, retention, and online optimization, and thus represents an important and closely related line of research on inference-time adaptation.
More broadly, we view Miras as a meaningful contribution toward adaptive and context-aware language modeling under distribution shift.

\paragraph{Why we do not include a direct controlled comparison.}
At the same time, a direct controlled comparison in our setting is not currently feasible.
To the best of our knowledge, Miras does not provide released code, pretrained model weights, or a reproducible training configuration, and no publicly accessible pretrained checkpoint at the 1.3B scale is available.
This limitation is particularly important in our setting, where fair comparison requires controlled pretraining, matched token budgets, and closely aligned backbone scales, especially for specialized-domain evaluation.
Therefore, we do not believe that an uncontrolled re-implementation would support a reliable apples-to-apples comparison.

\paragraph{Conceptual difference.}
More importantly, the differences between Nirvana and Miras are not merely implementation-level variations, but reflect distinct design choices that materially affect adaptation behavior, efficiency, and robustness.
First, Nirvana adopts a \emph{weight-bank structured} fast-parameter design, whereas Miras does not explicitly impose such structure on fast parameterization.
This design allows Nirvana to reuse and interpolate a shared set of basis blocks, which improves parameter efficiency and makes fast-weight dynamics more controlled.
Second, Nirvana uses a \emph{task-conditioned} updating mechanism: the Trigger extracts task-dependent signals and modulates fast-parameter evolution through CL-OGD and conditional interpolation.
By contrast, Miras is not built around the same form of task-conditioned flexible updater.
Third, Nirvana explicitly evaluates adaptation under \emph{specialized-domain shift}, including Biomed, Finance, Law, and MRI, which provides a more direct test of robustness and practical adaptability beyond standard general-domain benchmarks.

These differences are substantive rather than cosmetic.
Taken together, weight-bank structuring, task-conditioned updating, and specialized-domain evaluation move Nirvana toward a setting where fast adaptation is not only effective, but also more stable, interpretable, and practically relevant under distribution shift.

\begin{table*}[t]
\centering
\begin{tabular}{lll}
\toprule
\textbf{Aspect} & \textbf{Miras \citep{behrouz2025s}} & \textbf{Nirvana} \\
\midrule
Fast parameterization & Not weight-bank structured & Weight-bank structured \\
Memory updater & Fixed mechanism & Flexible task-related mechanism \\
Specialized-domain evaluation & Not included & Biomed, Finance, Law, MRI \\
\bottomrule
\end{tabular}
\caption{Comparison between Miras and Nirvana.}
\label{tab:miras_vs_nirvana}
\end{table*}

\section{Experiments Related to RoPE}
\label{appdix_rope}

We investigate whether adding RoPE \citep{rope} in SWA enhances the model's capability or not, where Nirvana-RoPE refers to the Nirvana model with RoPE added to the query and key in SWA modules. 
We first conduct experiments on NIAH in Table \ref{niah2}.
Note that if RoPE is added to the SWA module, Nirvana-RoPE will be drastically worse than Nirvana  without RoPE in 8K context length, with an accuracy of only 0.2\% on S-NIAH-PK and an accuracy of only 4.4\% on S-NIAH-N. 
This illustrates the importance of removing RoPE in SWA modules, which can lead to degraded performance when the context length at the test time is larger than that at the training time.

\begin{table*}[t]
  \centering
  \resizebox{\textwidth}{!}{
  \begin{tabular}{l|ccc|ccc|ccc|ccc|cc|c}
  \hline
  \multirow{2}{*}{Models} &
  \multicolumn{3}{c|}{Single-Doc QA} &
  \multicolumn{3}{c|}{Multi-Doc QA} &
  \multicolumn{3}{c|}{Summarization} &
  \multicolumn{3}{c|}{Few-shot} &
  \multicolumn{2}{c|}{Code} &
  \multirow{2}{*}{Avg} \\
  & NQA & QQA & MFQ & HQA & 2WM & Mus & GvR & QMS & MNs & TRC & TQA & SSM & LCC & RBP & \\
  \hline
  Recurrent models &  &  &  &  &  &  &   &  &  &  &  &  &  &   \\
  RetNet & 12.1 & 10.7 & 19.1 & 10.7 & 18.0 & 5.8 & 4.8 & 15.8 & 7.9 & 19.0 & 18.0 & 12.8 & 14.1 & 17.9 & 13.2 \\
  HGRN2 & 10.7 & 12.1 & 19.1 & 11.3 & 15.7 & 6.0 & 5.2 & 15.1 & 9.2 & 16.0 & 15.8 & 10.3 & 18.6 & 20.8 & 13.5 \\
  Mamba & 13.0 & 10.1 & 20.4 & 10.1 & 16.7 & 6.0 & 7.2 & 15.9 & 8.4 & 23.1 & 21.9 & 11.2 & 17.9 & 19.0 & 14.6 \\
  DeltaNet & 12.9 & 10.8 & 21.5 & 10.9 & 13.2 & 5.1 & 6.5 & 13.5 & 7.2 & 15.5 & 23.3 & 11.6 & 17.6 & 20.3 & 13.6 \\
  Mamba2 & 11.1 & 11.3 & 18.6 & 11.8 & 15.1 & 6.7 & 6.7 & 14.5 & 7.4 & 13.0 & 23.6 & 8.4 & 17.9 & 20.6 & 13.5 \\
  Gated DeltaNet & 14.1 & 14.0 & 23.3 & 13.7 & 14.4 & 5.8 & 7.5 & 16.4 & 7.9 & 30.0 & 22.4 & 23.0 & 18.7 & 22.1 & 16.6 \\
  \hline
  Attention or hybrid models &  &  &  &  &  &  &   &  &  &  &  &  &  &   \\
  Transformer++     & 11.8 & 9.3  & 10.0 & 10.9 & 4.2  & 6.1 & 7.4 & 15.8 & 6.6  & 16.9 & 13.5 & 3.9 & 17.2 & 18.7 & 11.0 \\
  Samba             & 12.5 & 12.9 & 25.4 & 11.2 & 19.7 & 6.8 & 9.1 & 15.7 & 11.0 & 20.0 & 22.7 & 22.8 & 18.1 & 21.1 & 15.9 \\
  Gated DeltaNet-H1 & 14.5 & 12.3 & 26.6 & 12.6 & 23.6 & 6.1 & 9.1 & 16.1 & 12.8 & 33.5 & 23.9 & 26.8 & 15.5 & 19.2 & 17.8 \\
  Gated DeltaNet-H2 & 12.7 & \textbf{13.0} & \textbf{27.1} & 12.7 & 20.6 & 7.5 & \textbf{10.4} & \textbf{16.2} & 13.0 & \textbf{40.5} & 22.7 & \textbf{27.9} & \textbf{19.9} & \textbf{22.1} & 18.4 \\
  Nirvana-noTrigger     & 14.8 & 11.8 & 25.6 & 14.0 & 23.9 & 7.7 & 9.2  & 15.1 & 13.5 & 33.0 & 21.2 & 22.9 & 16.5 & 20.9 & 17.9 \\
  \textbf{Nirvana (Ours)}           & \textbf{16.6} & 12.8 & 26.0 & \textbf{14.6} & \textbf{24.8} & \textbf{9.7} & \textbf{10.4} & 15.9 & \textbf{15.4} & 36.4 & \textbf{25.2 }& 22.6 & 17.5 & 21.5 & \textbf{19.2} \\
  \hline
  \end{tabular}
  }
      \caption{Accuracy on 14 tasks from LongBench \citep{bai2023longbench}, including Narrative QA, QasperQA, MultiField QA, HotpotQA, 2WikiMulti QA, Musique, GovReport, QMSum, MultiNews, TRec, Trivia QA, SamSum, LCC, and RepoBench-P by order.}
  \label{longbench_table}
  \end{table*}

We conduct experiments on language modeling and zero-shot common sense reasoning in Table~\ref{arc2}.
The results illustrate that adding RoPE does not make Nirvana's performance better on average accuracy.
The reason is that the linear attention architectures (e.g., Gated DeltaNet and Mamba2) are well qualified to capture the position-dependent information of the input sequence \citep{yang2025gated}. 
Thus, there is no need to use RoPE in SWA, which would otherwise require additional computation and undermine the model's ability to extrapolate with context lengths longer than the training data.


\section{Supplementary Language Modeling Ability} 

We evaluate model performance on LongBench \citep{bai2023longbench}, a comprehensive suite of long-context tasks spanning retrieval, reasoning, multi-document understanding, and in-context learning. As shown in Table~\ref{longbench_table}, Nirvana achieves consistent improvements across most categories—including NQA, HQA, 2WM, Mus, GvR, MNs, and TQA. These results highlight Nirvana’s strengthened abilities in long-range retrieval, efficient in-context learning, and robust state tracking, demonstrating its effectiveness not only in general domains but also in specialized long-context understanding.

\begin{figure*}[t]
    \centering
    \begin{subfigure}[t]{0.32\linewidth}
      \centering
      \includegraphics[width=\linewidth]{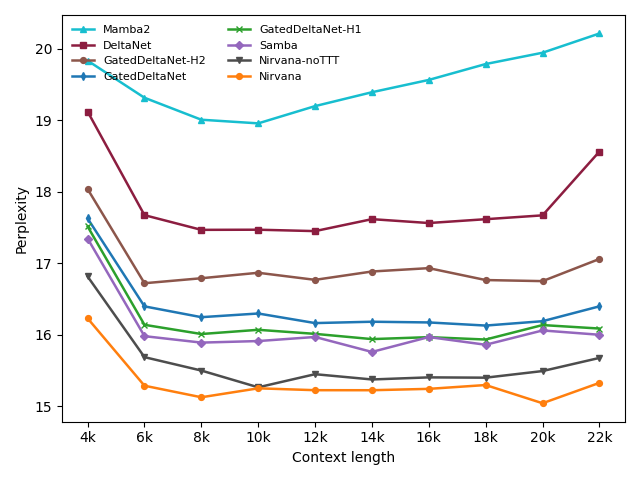}
      \subcaption{NarrativeQA}
    \end{subfigure}
    \hfill
    \begin{subfigure}[t]{0.32\linewidth}
      \centering
      \includegraphics[width=\linewidth]{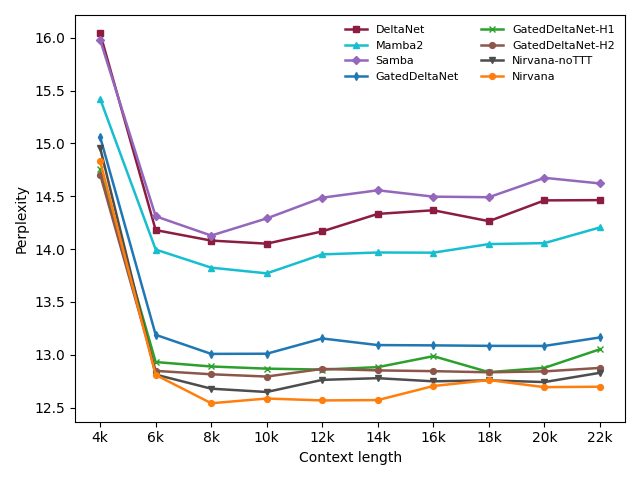}
      \subcaption{QMSum}
    \end{subfigure}
    \hfill
    \begin{subfigure}[t]{0.32\linewidth}
      \centering 
      \includegraphics[width=\linewidth]{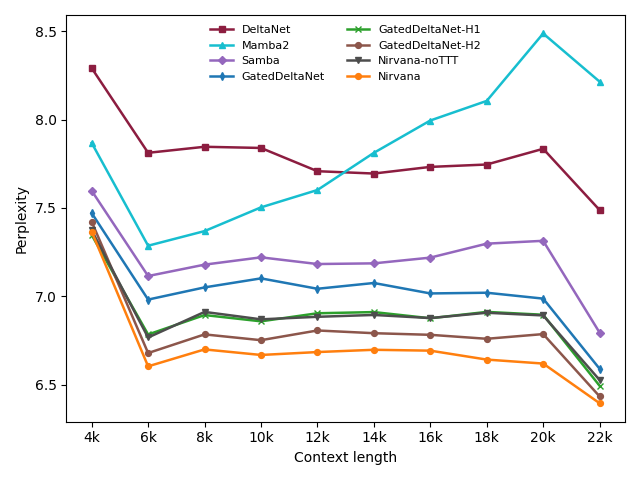}
      \subcaption{GovReport}
    \end{subfigure}
    \caption{Length extrapolation from 4K to 20K tokens on three long benchmarks.}
    \label{long_extropolation}
  \end{figure*}
  
As shown in Figure~\ref{long_extropolation}, 
we evaluate the models' capacity of extrapolating to sequences from 4K to 20K tokens
across three long-context benchmarks, i.e., NarrativeQA, QMSum, and GovReport \citep{bai2023longbench}. 
Nirvana achieves the lowest overall perplexity across different tasks among all models. 
Besides, Nirvana without Trigger is also evaluated in Figure~\ref{long_extropolation}, and its performance is not as good as that of Nirvana. 
While we observe performance fluctuations when the context length becomes longer, Nirvana exhibits relatively more stable performance, which indicates that Nirvana is robust and has superiority in length-extrapolation tasks. 
We will explore Nirvana's capabilities on even longer sequences in the future.  

\begin{table*}[t]
\centering
\begin{tabular}{l|ccccccc}
\hline
Models            & SWDE & SQD & FDA & TQA & NQ & Drop & Avg \\
\hline
Recurrent models  &      &      &     &     &    &       &     \\
RetNet            & 14.0 & 28.5 & 7.0 & 54.4 & 16.2 & 17.3 & 22.9 \\
HGRN2             &  8.3 & 25.3 & 4.8 & 51.2 & 14.2 & 16.9 & 20.1 \\
Mamba             &  9.8 & 25.8 & 3.7 & 54.3 & 14.9 & 17.4 & 21.0 \\
Mamba2            & 19.1 & 33.6 & 25.3 & 61.0 & 20.8 & 19.2 & 29.8 \\
DeltaNet          & 17.9 & 30.9 & 18.4 & 53.9 & 17.3 & 18.6 & 26.2 \\
Gated DeltaNet    & 25.4 & 34.8 & 23.7 & 60.0 & 20.0 & 19.8 & 30.6 \\
\hline
Attention or hybrid models &  &  &  &  &  &  &  \\
Transformer++     & 29.5 & 38.0 & 52.2 & 58.3 & 22.5 & 21.6 & 37.0 \\
Samba             & 33.0 & 39.2 & 50.5 & 57.7 & 23.5 & 20.2 & 37.3 \\
Gated DeltaNet-H1 & 35.6 & 39.7 & \textbf{52.0} & 60.1 & 24.6 & 22.2 & 39.0 \\
Gated DeltaNet-H2 & \textbf{38.2} & 40.4 & 50.7 & \textbf{63.3} & \textbf{24.8} & \textbf{23.3} & \textbf{40.1} \\
Nirvana-noTrigger     & 35.1 & 39.8 & 50.5 & 60.0 & 22.3 & 21.7 & 38.2 \\
\textbf{Nirvana (Ours)}  & 37.8 & \textbf{41.0} & 51.1 & 62.8 & \textbf{24.8} & 22.9 & \textbf{40.1} \\
\hline
\end{tabular}
\caption{Accuracy on recall-world retrieval tasks with the input sequences truncated to 2K tokens, where SQD is short for SQUADE, and TQA is short for Trivial QA \citep{icl}.}
\label{icl}
\end{table*}

In Table~\ref{icl}, we present the models' accuracy on real-world recall-intensive tasks \citep{icl}.
Due to the limitations of linear attention, recurrent models show a significant performance gap compared to Transformers++, 
while Nirvana outperforms Transformers++ and achieves comparable performance with SOTA hybrid models in retrieval-intensive tasks. 
Without Trigger, Nirvana's performance will be notably degraded because of the lack of crucial task-aware memory management mechanism. 

\section{Specialized Ability of MRI Reconstruction}
\label{appdix_mri}

\begin{table}[t]
  \centering
  \resizebox{\columnwidth}{!}{%
\begin{tabular}{lcc}
  \hline Undersampling Rate & Acceleration Rate & Center Fraction Rate \\
  \hline 
  $12 \times$ & 12 & 0.027 \\
  $10 \times$ & 10 & 0.032 \\
  $8  \times$ & 8  & 0.04 \\
  $6  \times$ & 6  & 0.06 \\
  $4  \times$ & 4  & 0.08 \\
  \hline
\end{tabular}
}
      \caption{The k-space undersampling configurations (acceleration and center fraction parameters) used for MRI reconstruction.}
        \label{undersampling}
\end{table}

In MRI reconstruction, we undersample the k-space signals to accelerate the MRI coil scanning process in the frequency domain, and at the same time also reduce the amount of data to be processed \citep{fastmri,varnet,udno}.
The detailed k-space undersampling configurations are shown in Table~\ref{undersampling}.

\begin{figure*}[t]
  \centering
  \begin{subfigure}[t]{0.32\linewidth}
    \centering
    \includegraphics[width=\linewidth]{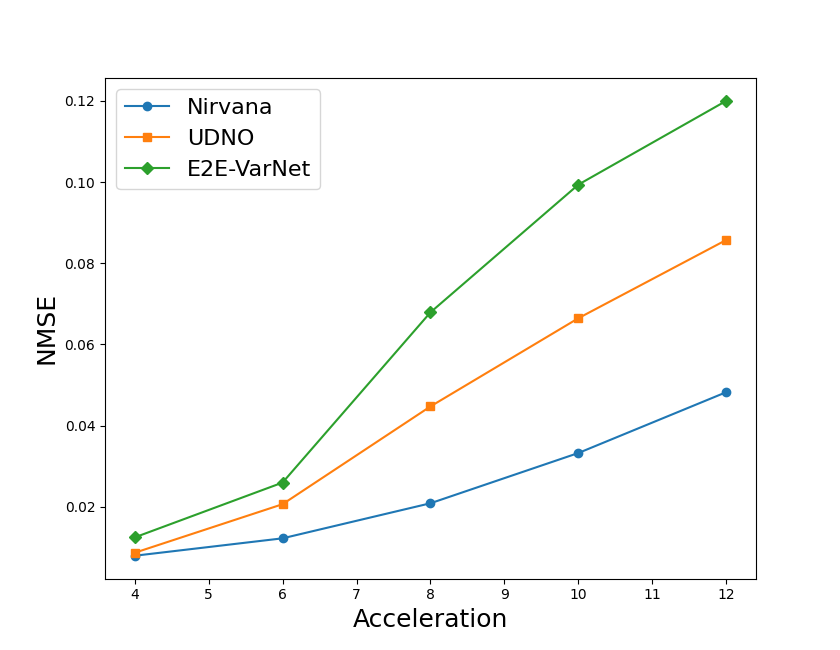}
    \subcaption{NMSE}
  \end{subfigure}
  \hfill
  \begin{subfigure}[t]{0.32\linewidth}
    \centering
    \includegraphics[width=\linewidth]{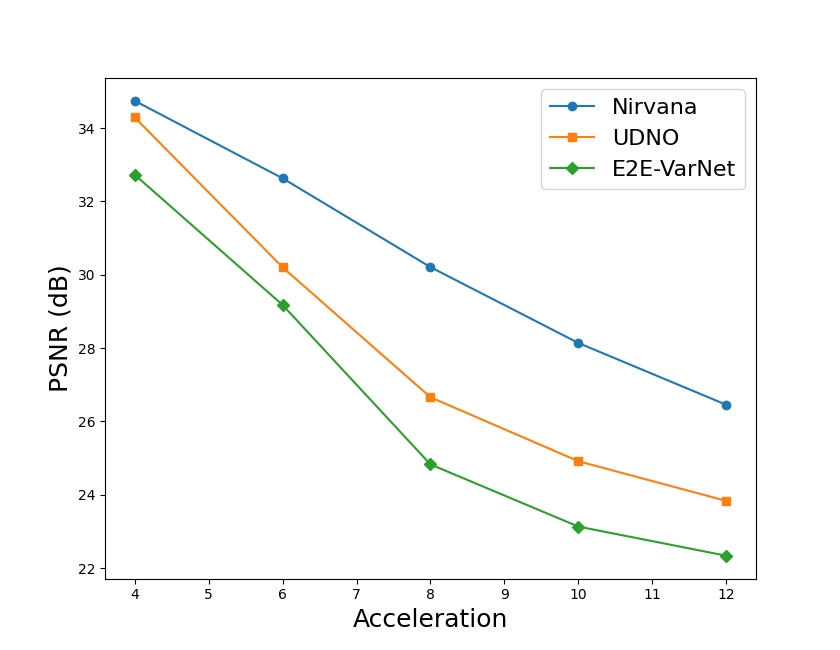}
    \subcaption{PSNR}
  \end{subfigure}
  \hfill
  \begin{subfigure}[t]{0.32\linewidth}
    \centering
    \includegraphics[width=\linewidth]{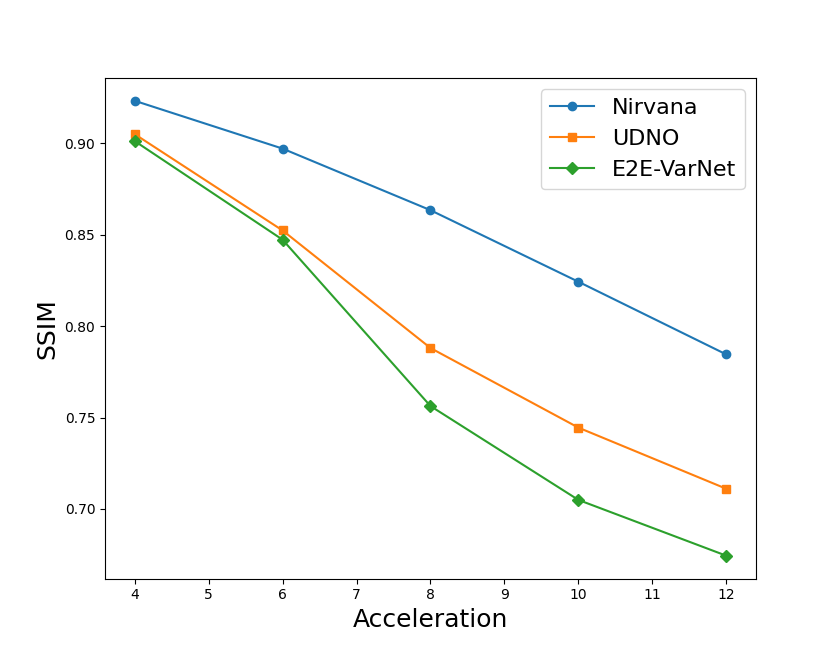}
    \subcaption{SSIM}
  \end{subfigure}
  \caption{MRI reconstruction performance comparison for models with 160M trainable parameters. The acceleration rate is also the undersampling rate.}
  \label{mri_plot}
\end{figure*}

We compare the performance of Nirvana for MRI reconstruction with other baselines under different undersampling rates in Figure~\ref{mri_plot}. 
The MRI reconstruction performances of all models degrade when the undersampling rate becomes larger, because less information is provided in the higher-rate undersampled k-space signals. 
Nirvana surpasses the other models under all undersampling rates, and the Nirvana's performance degradation trend is the least significant as the undersampling rate becomes larger. 
This illustrates the potential advantage of Nirvana, which can use highly undersampled k-space signals to reconstruct the image while maintaining the same or even better image quality compared to E2E-VarNet and UDNO. 
Therefore, Nirvana has the potential ability to accelerate the scanning process of MRI.

\begin{table}[ht]
  \centering
    \resizebox{\columnwidth}{!}{%
  \begin{tabular}{l | ccc}
    \toprule
    Model & SSIM $\uparrow$ & PSNR (dB) $\uparrow$ & NMSE ($\times 10^{-2}$) $\downarrow$ \\
    \midrule
    E2E-VarNet                 & 0.8540 ± 0.0418 & 29.68  ± 2.99 & 2.512 ± 0.742 \\
    UDNO                       & 0.8598 ± 0.0414 & 30.21  ± 2.97 & 2.074 ± 0.730 \\
    \textbf{Nirvana (Ours)}    & \textbf{0.9003} ± 0.0407 & \textbf{32.97}  ± 2.93 & \textbf{1.176} ± 0.625 \\ 
    \bottomrule
  \end{tabular}
  }
      \caption{MRI reconstruction performance comparison of models with 160M trainable parameters. 
  For Nirvana, the trainable components are the k-space encoder and the MRI decoder. 
  The undersampling rate is set as 6 in this table during the test time.}
    \label{mri}
\end{table}

In Table~\ref{mri}, we evaluate the performance of Nirvana for MRI reconstruction using SSIM, PSNR, and NMSE, and also compare its performance with other baselines, including E2E-VarNet \citep{varnet} and UDNO \citep{udno}. 
The undersampling rate is set as 6 in the test time. 
As shown in Table~\ref{mri}, Nirvana achieves the highest SSIM and PSNR, 
as well as the lowest NMSE on the test set.
Besides, Nirvana's performance has the smallest variance and thus the highest stability. 
Specifically, Nirvana achieves an average improvement of 0.0405 in SSIM, 2.76 dB in PSNR, and $8.974 \times 10^{-3}$ in NMSE compared to the SOTA model UDNO \citep{udno}, respectively. 

\begin{figure*}[t]
  \centering
  \begin{tabular}{cccc}
    \textbf{Ground Truth} & \textbf{E2E-VarNet} & \textbf{UDNO} & \textbf{Nirvana (Ours)} \\
    \includegraphics[width=0.22\textwidth]{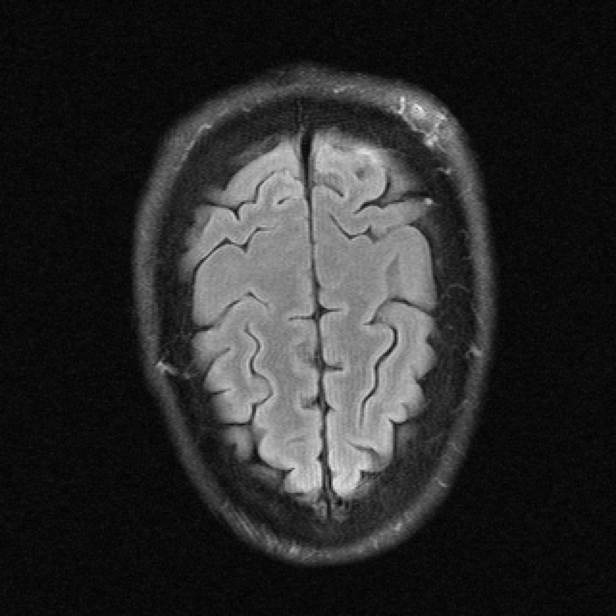} &
    \includegraphics[width=0.22\textwidth]{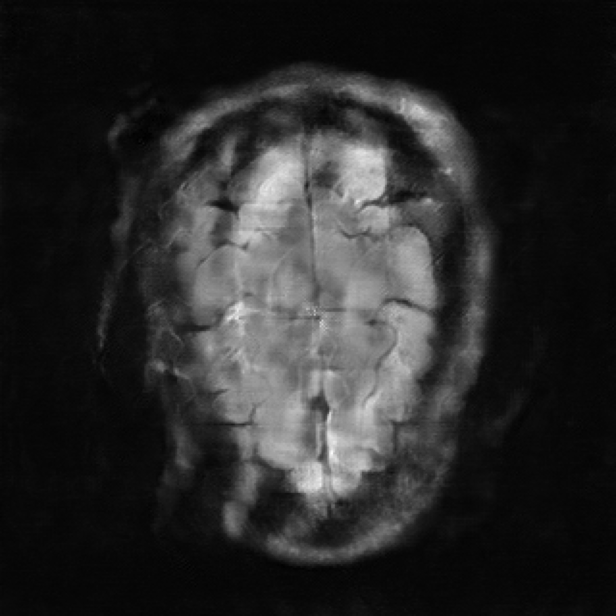} &
    \includegraphics[width=0.22\textwidth]{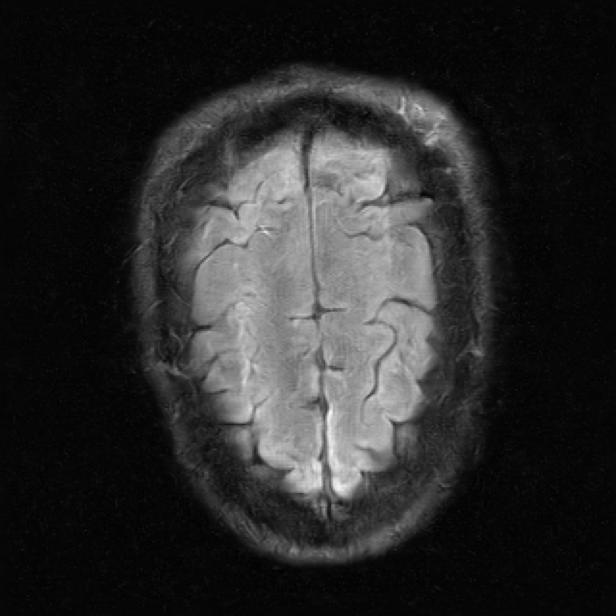} &
    \includegraphics[width=0.22\textwidth]{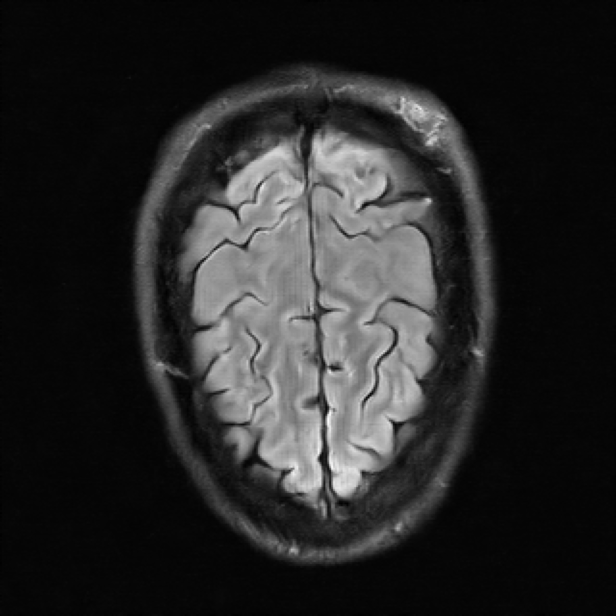} \\
  \end{tabular}
  \caption{MRI reconstruction performance comparison for models with 160M trainable parameters. The acceleration rate, i.e., the undersampling rate, is set as 8 in the test time. 
}
  \label{mri_img}
\end{figure*}
 
We further visualize Nirvana’s MRI reconstruction performance at an undersampling rate of 8 in Figure~\ref{mri_img}.
The ground truth, the images reconstructed by E2E-VarNet, UDNO, and Nirvana are shown in the 4 columns, respectively. 
As shown in Figure~\ref{mri_img}, the performance of Nirvana is better than UDNO and E2E-VarNet in terms of the image fidelity and resolution. 
The reconstructed image of E2E-VarNet is blurry, and some part of the brain is completely obscured by black patches. 
The reconstructed image of UDNO is roughly close to the ground truth image, but the resolution is low and the details of the image are unclear.
However, Nirvana delivers the clearest and most accurate high-resolution reconstruction, showing the closest resemblance to the ground truth and attaining the highest SSIM of 0.8812.

\section{A Toy Example of Nirvana in Combinatorial Tasks}

\begin{figure}[t]
\centering
\centerline{\includegraphics[width=7.9cm]{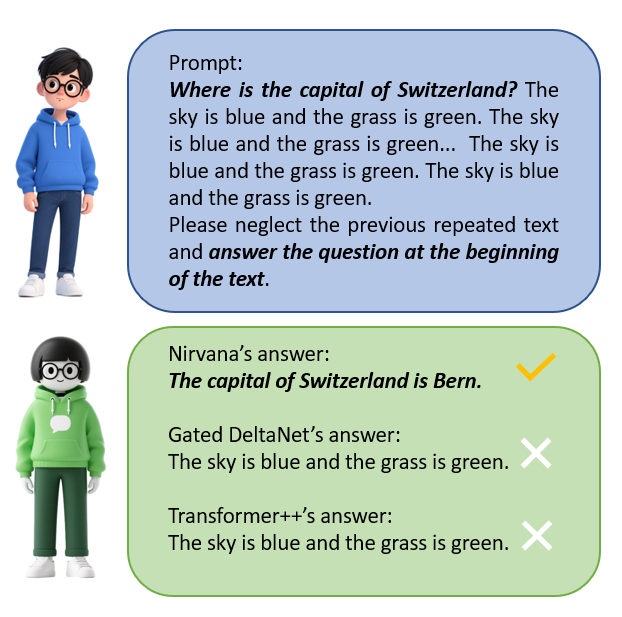}} 
\caption{A toy example for combinatorial tasks of common sense reasoning and key information retrieval in long sequences. } 
\label{toy_example}
\end{figure}

To illustrate the effectiveness of Nirvana model, we consider a toy example of combinatorial tasks, where the model is required to conduct common sense reasoning while retrieving the question from a haystack.
As shown in Figure~\ref{toy_example}, the haystack contains a set of repeated useless information, such as "the sky is blue" and "the grass is green". The key information, i.e., the question, is "Where is the capital of Switzerland?" The model should be able to retrieve the useful information at the beginning of the haystack and then answer the question. 
The Nirvana model accurately distinguishes the useful question from the useless information and then answers the question correctly. 
However, both Transformer++ and Gated DeltaNet fail to find the question and are misled to repeat the useless message instead.
This demonstrates the superior performance of the Nirvana model over Transformer++ and Gated DeltaNet in combinatorial tasks of common sense reasoning and key information retrieval in long sequences.

\end{document}